\newif\ifconfver
\confvertrue        
\documentclass[10pt,journal,compsoc]{IEEEtran}

\ifCLASSOPTIONcompsoc

  \usepackage[nocompress]{cite}
\else
  \usepackage{cite}
\fi

\usepackage{calc,amsfonts,amssymb,amsmath,bm,url,color,theorem,graphicx,cite}
\usepackage{psfrag,subfigure,float}
\usepackage{multirow,subfigure,amsmath,epsfig,amsfonts,amssymb,psfig,graphics,psfrag,theorem,calc,subfigure,url,bm,cite}
\usepackage{stfloats}  
\usepackage{algorithm,algorithmic}
\usepackage{diagbox} 
\usepackage{hhline}
\usepackage{caption}
\usepackage{booktabs}
\usepackage{mathtools}

\def\blue{\color{blue}}
\def\red{\color{red}}

\usepackage[table,xcdraw]{xcolor}
\usepackage{colortbl} 

\makeatletter
\def\multilimits@{\bgroup
	\Let@
	\restore@math@cr
	\default@tag
	\baselineskip\fontdimen10 \scriptfont\tw@
	\advance\baselineskip\fontdimen12 \scriptfont\tw@
	\lineskip\thr@@\fontdimen8 \scriptfont\thr@@
	\lineskiplimit\lineskip
	\vbox\bgroup\ialign\bgroup\hfil$\m@th\scriptstyle{##}$\hfil\crcr}
\def\Sb{_\multilimits@}
\def\endSb{\crcr\egroup\egroup\egroup}
\makeatother

\definecolor{orange}{RGB}{255,107,0}
\def\blue{\color{blue}}
\def\red{\color{red}}

{\begin{list}{}%
		{\setlength{\rightmargin}{0pc}%
			\setlength{\leftmargin}{1pc} }} %
	{\end{list}}

\newlength{\twidth}
\ifconfver
\else
\fi

\definecolor{mediumseagreen}{rgb}{0.58, 0.44, 0.86}
\definecolor{orange}{RGB}{255,107,0}
\def\blue{\color{blue}}

\def\red{\color{red}}

\theorembodyfont{\rmfamily}

{\begin{list}{}{
			\settowidth{\labelwidth}{\mbox{\textnormal{#1}}}%
			\setlength{\leftmargin}{\labelwidth+\labelsep}}}%
	{\end{list}}

\newcommand\bW{\ensuremath{{\bm W}}}

\newcommand\bx{\ensuremath{{\bm x}}}

\newcommand\bZ{\ensuremath{{\bm Z}}}

\newcommand\bA{\ensuremath{{\bm A}}}

  \usepackage[nocompress]{cite}
  \usepackage{cite}
\usepackage{graphicx}
\usepackage{amsmath}
\usepackage{amssymb}
\usepackage{booktabs}
\usepackage{bbding}
\usepackage{epsfig}
\usepackage{multirow}
\usepackage{threeparttable}  
\usepackage{color}
\usepackage{array}
\usepackage{makecell}
\usepackage{stfloats}
\usepackage[table]{xcolor}
\usepackage[misc]{ifsym}
\usepackage[accsupp]{axessibility}
\usepackage[pagebackref,breaklinks,colorlinks]{hyperref}

\ifCLASSINFOpdf
\else
\fi

\begin{document}

\title{GRACE: Graph-Regularized Attentive Convolutional Entanglement with Laplacian Smoothing for Robust DeepFake Video Detection

	\thanks{This study was supported in part by the Ministry of Science and Technology (MOST), Taiwan, under grants MOST XXX; 
and partly by the Higher Education Sprout Project of Ministry of Education (MOE) to the Headquarters of University Advancement at National Cheng Kung University (NCKU).}
\thanks{\textit{(Corresponding author: Chih-Chung Hsu.)}}
\thanks{
		C.-C. Hsu, S.-N. Chen, M.-H. Wu, Y.-F. Wang, C.-M. Lee and Y.-S. Chou are with Institute of Data Science and Department of Statistics, National Cheng Kung University, Tainan, Taiwan (R.O.C.), (e-mail:cchsu@gs.ncku.edu.tw, johnnychen1999@gmail.com, re6091054@gs.ncku.edu.tw, re6113018@gs.ncku.edu.tw, zuw408421476@gmail.com, nelly910421@gmail.com.)}
}

\author{Chih-Chung Hsu,~\IEEEmembership{Senior Member,~IEEE}, Shao-Ning Chen, Mei-Hsuan Wu,\\ Yi-Fang Wang, Chia-Ming Lee, Yi-Shiuan Chou
}


\IEEEtitleabstractindextext{%
\begin{abstract}

As DeepFake video manipulation techniques escalate, posing profound threats, the urgent need to develop efficient detection strategies is underscored. However, one particular issue lies with facial images being mis-detected, often originating from degraded videos or adversarial attacks, leading to unexpected temporal artifacts on spatial domain that can undermine the efficacy of DeepFake video detection techniques. This paper introduces a novel method for robust DeepFake video detection, harnessing the power of the proposed Graph-Regularized Attentive Convolutional Entanglement (GRACE) based on the graph convolutional network to address the aforementioned challenges. First, conventional convolution neural networks are deployed to perform spatiotemporal features for the entire video. Then, the spatial and temporal features are mutually entangled by constructing a graph with sparsity constraint, enforcing essential features of valid face images in the noisy face sequences remaining, thus augmenting stability and performance for DeepFake video detection. Furthermore, the Graph Laplacian smoothing prior regularization is proposed in the graph convolutional network to suppress the noise pattern in the feature space to further improve the performance and robustness. Comprehensive experiments are conducted to illustrate that our proposed method delivers state-of-the-art performance in DeepFake video detection under noisy and unreliable face sequences. 
The source code is available at \hyperlink{https://github.com/ming053l/GRACE}{https://github.com/ming053l/GRACE}.

\end{abstract}

\begin{IEEEkeywords}
		DeepFake Detection, Feature Entanglement, Graph Convolution Network, Adversarial Attack, Forgery Detection.
\end{IEEEkeywords}}

\maketitle

\IEEEdisplaynontitleabstractindextext

\IEEEpeerreviewmaketitle

\ifCLASSOPTIONcompsoc
\IEEEraisesectionheading{\section{Introduction}\label{sec:introduction}}
\else
\section{Introduction}
\label{sec1}
\fi

\maketitle

\IEEEPARstart{W}{ith} the widespread use of fake images and videos on various social network platforms for creating fake news and defrauding personal information, identifying synthesized content generated by generative adversarial networks (GANs) and variational autoencoders (VAEs) has become a critical challenge. As generative models advance and improve rapidly, current DeepFake detection techniques struggle to maintain effectiveness. To address this, several large-scale fake image datasets, such as FaceForensics++ (FF++) \cite{ffplus}, DeepFake Challenge Dataset (DFCD) \cite{dfdc}, Celeb-DF \cite{celeb}, and WildDeepFake \cite{zi2020wildDeepFake} have been established to promote the development of effective DeepFake detection techniques.

DeepFake image and video manipulation techniques have emerged as the most well-known forgery generation applications, with far-reaching impacts on numerous individuals. Generally, facial manipulation schemes can be classified into four categories: 1) entire face synthesis, 2) attribute manipulation, 3) identity swap, and 4) expression swap. Identity swap schemes have the most significant impact as they can be used to fabricate fake news targeting specific politicians. Many DeepFake detection techniques focus on identifying such fake videos using supervised learning methods with a pre-collected large-scale training set \cite{afchar2018mesonet, chollet2017xception, fake1, fake2, wang2020cnn, nguyen2019multi, kim2021fretal, nguyen2019use, FTCN, ICT}.

All of the DeepFake video detection models, however, often assume that the input facial sequence is reliable and well-detected, as the current state-of-the-art face detectors show promising performance. A promising strategy to prevent manipulated faces from being detected by DeepFake video detectors, could be making adversarial examples for face detectors since these are the first pipelines for all DeepFake detection techniques. Numerous studies have demonstrated the effectiveness of adversarial attacks on face detectors \cite{bose2018adversarial}. Several recent adversarial perturbation strategies \cite{goodfellow2014explaining} \cite{carlini2017towards} \cite{baluja2017adversarial} \cite{xie2017adversarial} have been proposed, potentially rendering face detectors ineffective. For instance, the methods introduced in \cite{bose2018adversarial} and \cite{attack1} indicate that the detection rate can decay to less than $10\%$, implying the $90\%$ facial images in a face sequence could be invalid. These perturbed DeepFake videos can yield noisy face sequences with many invalid facial images, leading to unintended temporal feature jittering in temporal-clue-aware methods \cite{LRnet}\cite{wang2022adt}\cite{fakefreq}. These temporal artifacts can significantly degrade their performance, while invalid facial images may also diminish the effectiveness of frame-level DeepFake video detection schemes \cite{chollet2017xception}\cite{fakexray} because the final decision of a video is based on the majority voting.

\begin{figure}
    \centering
	\includegraphics[width=0.48\textwidth]{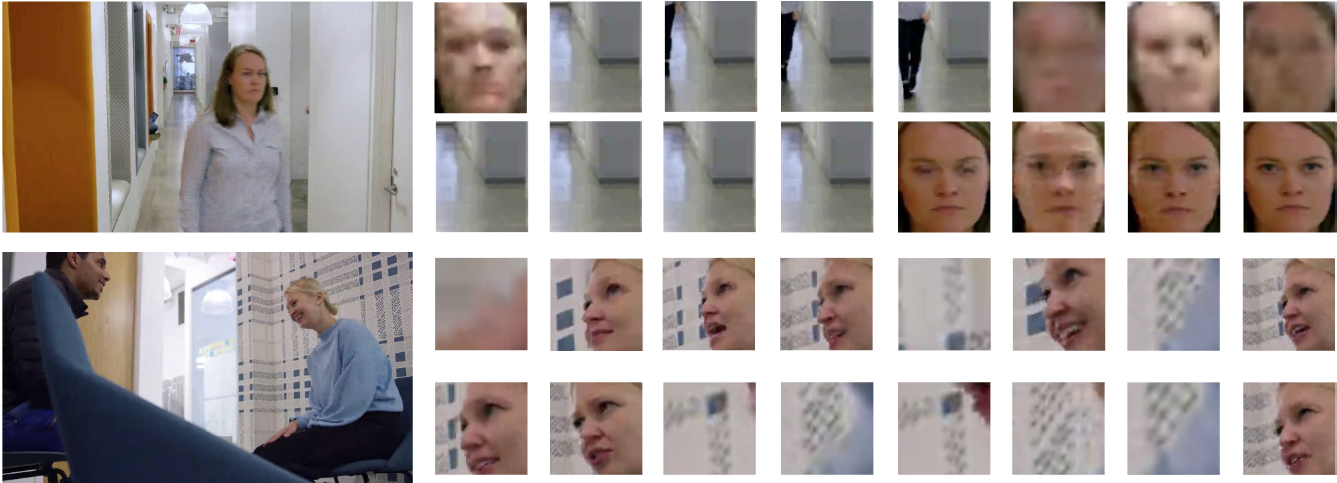}
	\caption{Example of the detected faces from two videos using RetinaFace (top) and Dlib (bottom). The sampled consecutive frames may not accurately detect faces, such noisy sequence may result in inconsistent spatial semantic features extracted by the feature extractor, which in turn makes practical application challenging.}
	\label{fig:fd}	
\end{figure}

\begin{table}[]
    \centering
    \caption{The confidence range of the detected faces using RetinaFace and Dlib for 200 videos sampled from FF++ \cite{ffplus}. Conf. and Det. stand for the confidence range of the detected faces using the specific face detector.}  
    \scalebox{1.08}{ 
    \begin{tabular}{c|ccc|c} 
        \hline\hline
        Det./Conf. & [0,0.33] & (0.33,0.66] & (0.66,0.1] & Total \\ \hline
        RetinaFace (raw) & 5 & 19 & 176 & 200 \\ 
        RetinaFace (c23) & 6 & 28 & 166 & 200 \\
        RetinaFace (c40) & 8 & 43 & 149 & 200 \\ \hline
        Dlib (raw) & 0 & 13 & 187 & 200 \\
        Dlib (c23) & 0 & 18 & 182 & 200 \\
        Dlib (c40) & 2 & 33 & 165 & 200 \\
        \hline\hline
    \end{tabular}
    \label{tab:fd_result}}
\end{table}

Even video compression could reduce the detection rate of the face detectors. We randomly select 200 videos from the FF++ dataset \cite{ffplus}, featuring varying compression ratios (raw, c23, c40), and extract 16 frames from each video for face detection analysis. We employ state-of-the-art face detection tools, such as RetinaFace \cite{deng2020retinaface} and Dlib \cite{dlib}, to substantiate our observations, as shown in the Figure \ref{fig:fd}. 
Table \ref{tab:fd_result} presents the face detection outcomes for the sampled videos. Notably, the predicted probability of 8 and 2 videos using RetinaFace \cite{deng2020retinaface} and Dlib \cite{dlib} falls within the $[0, 0.33]$ range, respectively, implying there are 8 and 2 facial images are mis-detected. Additionally, for uncompressed videos, 19 videos exhibit accuracy lower than $66\%$, highlighting the imperfections of face detectors.  The question is raised: \textbf{\textit{Could the current DeepFake detection methods be robust to such noisy face sequences?}} The answer is negative. We simply replace the $40\%$ facial images with background ones for the testing set of FF++ \cite{ffplus} with raw setting and evaluate the performance using Xception \cite{chollet2017xception}. Unsurprisingly, the accuracy dropped significantly after replacement. An effective solution to deal with the issues raised by noisy face sequences for DeepFake video detection is highly desired.

\begin{figure*}
		\centering
		\includegraphics[width=1\textwidth]{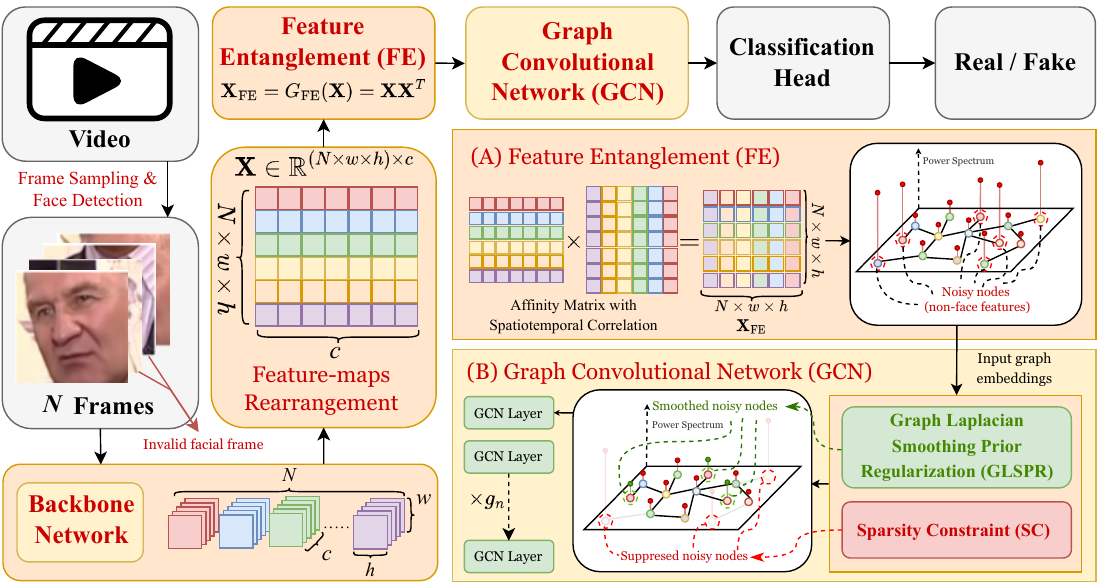}
		\caption{Flowchart of the proposed GRACE for robust DeepFake video detection. In practice, face detection results are not always effective. Invalid faces may significantly outnumber valid faces, limiting the DeepFake detection methods. We leverage the proposed Feature Entanglement technique to embed the spatio-temporal features. Graph Laplacian Smoothing Prior Regularization and Sparsity Constraint to smooth and suppress these noisy nodes.}
		\label{fig:features}	
\end{figure*}

In light of the escalating threat posed by various malicious attacks on face detectors that aim to undermine their reliability, this paper presents a pioneering Graph-Regularized Attentive Convolutional Entanglement (GRACE) with Laplacian Smoothing learning approach. GRACE leverages contextual features in both spatial and temporal domains to effectively detect DeepFake videos under noisy face sequences. By employing the proposed Feature Entanglement (FE) technique, an affinity matrix is constructed to amalgamate the spatiotemporal features, ensuring that each node possesses at least one feature descriptor originating from valid face images. Ultimately, Graph Laplacian Smoothing Prior regularization (GLSPR) is ingeniously integrated into the Graph Convolutional Network (GCN) to further suppress noisy nodes. Meanwhile, we meticulously incorporate Sparsity Constraint (SC) into network optimization term to prioritize the features of valid face images within the noisy face sequence, thereby significantly enhancing the performance of DeepFake video detection. The main contributions of this paper are three-fold:
\begin{itemize}
\item We propose a novel GRACE, which exploits contextual features in both spatial and temporal domains for robust DeepFake video detection under noisy face sequences. To the best of our knowledge, this is the first work to address the issue of unreliable face sequences for DeepFake video detection.

\item We introduce a FE mechanism to construct an affinity matrix that mixes the spatiotemporal features together, ensuring the efficacious representation learning of GCN. This approach effectively mitigates the impact of invalid facial images in the noisy face sequence.

\item We propose a GLSPR and SC to filter the noisy nodes further and improve the performance of DeepFake video detection. Comprehensive experiments demonstrate that our method achieves state-of-the-art performance, especially under challenging scenarios with unreliable and noisy face sequences.
\end{itemize}

    The rest of this paper is organized as follows. Section \ref{sec:relatedwork} introduces the related works in DeepFake detection. Section \ref{sec:proposed_method} presents the proposed GRACE architecture design. In Section \ref{sec:experiments}, the superiority of GRACE over benchmark methods is experimentally demonstrated. Finally, conclusions are drawn in Section \ref{sec:conclusions}.

\section{Related Works}\label{sec:relatedwork}


\subsection{General DeepFake Video Detection}\label{sec:generalrelatedwork}
In the domain of DeepFake video detection, numerous sophisticated approaches have recently emerged \cite{masi2020two}\cite{fakevideo1}\cite{fakexray}\cite{wang2022adt}\cite{LRnet}\cite{sabir2019recurrent} \cite{yang2019exposing}\cite{li2018ictu}\cite{fakecatcher}\cite{fakewarping}. On the one hand, these methods extend DeepFake image detection techniques by averaging the predictions of individual frames to assess a video's authenticity \cite{fakexray}\cite{li2018ictu}\cite{fakecatcher}\cite{fakewarping}. On the other hand, the temporal inconsistency feature is exploited for DeepFake video classification using supervised learning approaches, as demonstrated in \cite{LRnet}\cite{sabir2019recurrent}\cite{masi2020two}\cite{fakevideo1}. Recently, several advanced techniques have been proposed to enhance DeepFake video detection performance. To address the generalizability issue, semi-supervised learning is considered in \cite{hsu2020deep, hsuicip, hsu2} to capture common fake features from selected representative GANs, assuming that most GANs might share similar identifiable clues. Pairwise learning is then employed to learn these common features from the training set, improving generalizability for DeepFake image detection \cite{hsuicip, hsu2020deep}. 
CORE \cite{ni2022core} introduces a novel approach for learning consistent representations across different frames, while RECCE \cite{RECCE} employs a reconstruction-classification learning scheme to capture more discriminative features. DFIL \cite{pan2023dfil} proposes an incremental learning framework that exploits domain-invariant forgery clues to improve generalization ability. TALL-Swin \cite{xu2023tall} utilizes a thumbnail layout and Swin Transformer to learn robust spatiotemporal features for DeepFake detection. UCF \cite{yan2023ucf} focuses on uncovering common features shared by different manipulation techniques to enhance generalizability. DFGaze \cite{DFGAZE} uses the gaze analysis to analyze the gaze features of face video frames, and then applies a spatial-temporal feature aggregator to realize authenticity classification.

\subsection{External Reference DeepFake Video Detection}\label{sec:referencerelatedwork}

As Face X-ray \cite{fakexray} demonstrates superior performance and robustness, recent DeepFake video detection techniques concentrate on uncovering more reliable signatures produced by GANs to enhance detection performance. Specifically, state-of-the-art DeepFake video detection primarily focuses on exploiting various priors, such as bio-informatics clues \cite{fakecatcher}, facial warping artifacts \cite{fakewarping}, and noise patterns \cite{fakexray}. 
FakeCatcher \cite{fakecatcher} utilizes a novel bio-feature,  Photoplethysmography (PPG) response, to differentiate DeepFake videos, as real and fake videos exhibit distinct PPG features. A critical limitation is the necessity for high-resolution videos and images to effectively capture PPG cues. Moreover, The study in \cite{fakewarping} investigates warping artifacts at the boundaries of DeepFake videos, which arise due to the limited resolution of synthesized facial components. Concurrently, The study in \cite{masi2020two} introduces a method utilized deep Laplacian of Gaussian (LoG) and the loss of isolated manipulated faces to bolster the generalizability of DeepFake video tasks.


\subsection{Robust DeepFake Video Detection}\label{sec:advrelatedwork}
Recent research has concentrated on developing robust DeepFake detection models for compressed videos, as seen in \cite{LRnet}\cite{fakefreq}\cite{qian2020thinking}. The frequency component analysis method is employed to uncover intrinsic features and enhance performance under compression settings \cite{fakefreq}. However, frequency-aware features may prove ineffective under high compression (with high-frequency reduction) or noisy conditions (with high-frequency amplification). The {$F^3$}-net \cite{qian2020thinking} selects two complementary frequency bands as clues, devising a novel network to learn frequency-aware features that reveal subtle forgery artifacts. 
Recent studies \cite{carlini2020evading}\cite{hussain2021adversarial}\cite{neekhara2021adversarial} have highlighted the vulnerability of DeepFake detectors to adversarial perturbations. Therefore, adversarial defense with DeepFake detection, such as \cite{cchsupami}, have been attracting recently. The study in \cite{cchsupami} demonstrates an effective solution by leveraging statistical inference on CNNs to achieve better robustness to adversarial examples.

\section{Proposed Graph-Regularized Attentive Convolutional Entanglement}\label{sec:proposed_method}

\subsection{Overview of the Proposed Method}
The proposed GRACE's flowchart is illustrated in Figure \ref{fig:features}, which is the video-based approach. Previous research based on frequency analysis methods \cite{qian2020thinking} has suggested that high-frequency information could be critical in DeepFake detection. However, in practical scenarios, the detection process can be complicated by adversarial attacks or rapid object movements, leading to erroneous face detections and resulting in network vulnerability.

First, a face detector extracts facial images from each video frame. A CNN-based backbone network extracts high-level semantic features from the spatial domain of the acquired facial parts. Using the extracted spatial features, the spatial and temporal representations at frame $n$ and location $(i,j)$ across all feature maps $\mathbf{X}\in\mathbb{R}^{d\times c}$ can be obtained for the face sequence, where $d=N\times h\times w$, potentially including partially invalid faces. This feature representation captures feature responses at frame $n$ and location $(i,j)$ across all frames, thereby integrating temporal information, as shown in Figure \ref{fig:features}.

To augment the correlation between the spatial and temporal feature representation $\mathbf{X}$ acquired in the previous step, we introduce a novel Feature Entanglement, denoted as $\mathbf{X}_\textrm{FE}=G_\textrm{FE}(\mathbf{X}) \in\mathbb{R}^{d\times d}$, which carefully embeds both temporal and spatial features into its graph representation by affinity matrix $\mathbf{X}_\textrm{FE}$ from original feature $\mathbf{X}$. In highly noisy face sequences, the number of invalid faces could be more than that of valid ones. Therefore, the essential features could be relatively fewer. To efficiently discern the importance of the graph representation $\mathbf{X}_\textrm{FE}$, we introduce the GCN to capture the contextual features between nodes (spatiotemporal features) in $\mathbf{X}$. To further remove the noisy nodes from the original $\mathbf{X}_{\text{FE}}$, GLSPR is judiciously adapted to each layer of the GCN for better performance under noisy face sequences. Besides, we introduce SC into the objective function to allow our GRACE method to focus on these essential features. Finally, a softmax classifier is connected to the outcome of the GCN to evaluate the authenticity of the supplied facial parts.

\subsection{Feature Entanglement}

Traditional CNNs serve as the backbone network to obtain the spatial feature representation for each frame of the video. Assuming that the size of the extracted feature map is $c\times h\times w$, the spatial feature representation of a specific video at location $(i,j)$ via the backbone network can be vectorized into $\mathbf{x}_{n,i,j} = [x_{n,i,j}^1, x_{n,i,j}^2,...,x_{n,i,j}^c] 
\in\mathbb{R}^{c\times 1}$, where $c$ is the number of channels in the extracted spatial feature map, $i=1,...,h$, $j=1,...,w$, and $n=1,...,N$, with $N$ being the total number of frames. Then, let $d=N\times h\times w$, we create the feature context $\mathbf{X} \in \mathbb{R}^{d \times c}$ based on location-wise feature concatenation, as follows: 

\begin{equation}
\begin{aligned}
\mathbf{X} = [&(\bx^{1}_{1,1,1},\ \bx^{2}_{1,1,1},\ ...,\ \bx^{c}_{1,1,1}); \\
&(\bx^{1}_{1,1,2},\ \bx^{2}_{1,1,2},\ ...,\ \bx^{c}_{1,1,2}); \\
&\vdots \\
&(\bx^{1}_{1,h,w},\ \bx^{2}_{1,h,w},\ ...,\ \bx^{c}_{1,h,w}); \\
&\vdots \\
&(\bx^{1}_{N,h,w},\ \bx^{2}_{N,h,w},\ ...,\ \bx^{c}_{N,h,w})],
\end{aligned}
\end{equation}
where $\mathbf{X}$ represents the spatiotemporal feature.

The joint spatial and temporal feature representations from $\mathbf{X}$ necessitates addressing potential inefficiencies linked to considerable distances between the $n$-th and $m$-th feature vectors, $\bx^n$ and $\bx^m$, especially when $m$ and $n$ are significantly apart. Directing adopting the Transformer \cite{videoTrans}\cite{vit} could lead to large computational complexity since the number of tokens could be large enough to have a long-range correlation.

To address these issues and efficiently exploit joint spatial and temporal feature representations, we propose a novel concept,\textit{ i.e.}, Feature Entanglement, which is exactly an affinity matrix, creating a graph representation that encapsulates the relationships between nodes—each node representing entangled spatial and temporal features based on edges. Every element (node) in $\mathbf{X_\textrm{FE}}$ intertwines spatial and temporal information, thereby facilitating GCN to learn the relationships among spatiotemporal features more effectively, in which the features of the first and last frames might still have a link. The feature entanglement $\mathbf{X}_\textrm{FE}$ is defined as follows:

\begin{equation}
\mathbf{X}_\text{FE} = G_\textrm{FE}(\mathbf{X})=\mathbf{X}\mathbf{X}^T,
\end{equation}
where $\mathbf{X}_\text{FE} \in \mathbb{R}^{d \times d}$ is the affinity matrix obtained through feature entanglement. Consider, for instance, the feature element $\mathbf{X}_{\text{FE}}({1,c})$, which is computed by taking the inner product of the first and the $c$-th row of $\mathbf{X}$. This element encapsulates the correlation between the spatial features at location $(1,1)$ across all frames and the spatial features at location $(h,w)$ in the first frame, as shown in Figure \ref{fig:features}. 

\subsection{GCN with Graph Laplacian Smoothing}

GCNs have emerged as a powerful tool for processing graph-structured data \cite{gcn1,gcn2}, making them a suitable choice for handling the graph embedding obtained through feature entanglement. However, the noise level in each node can vary depending on the degree of distortion. For highly distorted or noisy nodes (i.e., those with entangled features primarily contributed by invalid facial images), it is beneficial to eliminate them to ensure stable and excellent performance in DeepFake video detection. To mitigate the impact of features from invalid facial images, we judiciously integrate the GLSPR, a well-established concept in Graph Signal Processing (GSP) \cite{gsp}, as a regularizer into the GCN to filter out highly noisy nodes \cite{gspla}.
%
%
In this study, we ingeniously leverage the properties of GLSPR from GSP and seamlessly incorporate it into the GCN framework, enabling end-to-end training without explicitly computing the eigendecomposition of the Graph Laplacian matrix (GLm).


\subsubsection{Graph Convolutional Network}

We construct affinity matrix $\mathbf{A} = \mathbf{X}_{\text{FE}} \in \mathbb{R}^{d \times d}$ through feature entanglement, where $\mathbf{A}_{ij}$ represents the edge weight between nodes $i$ and $j$. This matrix captures the similarity between different nodes. The degree matrix $\mathbf{D} \in \mathbb{R}^{d \times d}$ is a diagonal matrix where $\mathbf{D}_{ii} = \sum_j \mathbf{A}_{ij}$ represents the sum of edge weights connected to each node.

To further remove the redundancy among nodes, we apply adaptively thresholding to $\mathbf{A}$ to filter out weak or irrelevant connections. For each sample $i$, we compute the mean value of its feature entanglement matrix $\mathbf{A}$ and keep only the elements that are greater than half of the mean value. The indices and values of these elements are then extracted to form the edge indices and edge weights of a sparse affinity matrix $\mathbf{A}^{(i)}$, as follows:
\begin{equation}
\mathbf{A}^{(i)}_{jk} =
\begin{cases}
\mathbf{X}_{\text{FE},jk}^{(i)}, & \text{if } \mathbf{X}_{\text{FE},jk}^{(i)} >q \times \overline{\mathbf{X}_{\text{FE}}^{(i)}} \\
0, & \text{otherwise}
\end{cases}
\end{equation}
where $\mathbf{A}^{(i)}_{jk}$ denotes the edge weight between nodes $j$ and $k$ in the adjacency matrix of sample $i$, and $q$ is the factor controlling how strict the node being filtered. In this study, $q=0.5$ for all experiments. This thresholding operation helps to focus on the most important connections and reduces the computational burden of the GCN.

The resulting sparse adjacency matrix $\mathbf{A}^{(i)}$, along with the node features $\mathbf{X}^{(i)}$, serve as the input to the GCN for learning the graph structure and node representations. By combining the FE matrix $\mathbf{X}_\text{FE}$ and the sparse adjacency matrix $\mathbf{A}$, our method leverages the advantages of both representations, capturing rich spatiotemporal correlations while focusing on the most informative connections for the effective graph representation learning.

The core operation of GCN in the $l$-th layer can be described as follows:

\begin{equation}
\mathbf{Z}^{(l+1)} = \sigma(\hat{\mathbf{D}}^{-\frac{1}{2}} \hat{\mathbf{A}} \hat{\mathbf{D}}^{-\frac{1}{2}} \mathbf{Z}^{(l)} \mathbf{W}^{(l)}),
\label{formula:GCNpropagation}
\end{equation}
where $\hat{\mathbf{A}} = \mathbf{A} + \mathbf{I}_d$ is the adjacency matrix with self-loops, $\hat{\mathbf{D}}_{ii} = \sum_{j=1}^d \hat{\mathbf{A}}_{ij}$ is the corresponding degree matrix, $\mathbf{W}^{(l)}$ is the weight matrix of the $l$-th layer, $\mathbf{Z}^{(l)}$ represents the node features at layer $l$, and $\sigma$ is the activation function. This equation describes the aggregation and transformation of node features based on the graph structure. 

\subsubsection{Graph Laplacian Smoothing Prior Regularization}

While GCNs demonstrate exceptional capability in modeling long-range dependencies through prior graph structures, their performance may be compromised when confronted with highly noisy scenarios, limiting their applicability in real-world situations. To address this challenge, we propose the incorporation of GLSPR into GCNs to enhance their robustness and improve their performance in the presence of significant noise.
A noisy face sequence, potentially containing mis-detected faces, can result in inconsistent feature responses among neighboring nodes, thereby increasing the high-frequency components in the graph representation. GLSPR is particularly well-suited for attenuating these high-frequency components, effectively reducing the impact of noisy face sequences on DeepFake detection.

Given an undirected graph $G=(V,E)$, where $V$ is the set of nodes and $E$ is the set of edges, the GLm $\mathbf{L}$ is defined as:

\begin{equation}
\mathbf{L} = \mathbf{D} - \mathbf{A},
\end{equation}
where $\mathbf{D}$ is the degree matrix and $\mathbf{A}$ is the adjacency matrix of the graph $G$. The degree matrix $\mathbf{D}$ is a diagonal matrix, where $\mathbf{D}_{ii}$ equals the degree of node $i$ in $G$.
%

As a real symmetric matrix, the effectiveness of the GLm $\hat{\mathbf{L}}$ as a low-pass filter can be understood by examining its eigendecomposition:

\begin{equation}
\mathbf{L} = \mathbf{U}\Lambda\mathbf{U}^T,
\end{equation}
where $\mathbf{U}$ is the matrix of eigenvectors and $\Lambda$ is the diagonal matrix of eigenvalues. The eigenvalues of $\mathbf{L}$ represent the frequencies of the graph signals, with smaller eigenvalues corresponding to lower frequencies and larger eigenvalues corresponding to higher frequencies. To elaborate, the high/low frequencies mentioned above differ from the conventional frequency-aware Deepfake detection methods \cite{qian2020thinking}. In GSP \cite{gsp}, the GLm is often used as a low-pass filter to smooth signals defined on graphs, effectively suppressing high-frequency noise while preserving low-frequency information \cite{gspla}. 


To integrate the GLSPR into the GCN propagation rule, we propose a modified version of the GLm that incorporates the graph's structural information and enables effective feature smoothing and noise suppression, thereby reducing the temporal artifacts or semantic inconsistency on the spatial domain resulting from invalid facial images. Our proposed GLm, denoted as $\hat{\mathbf{L}}$, is defined as follows:

\begin{equation}
\hat{\mathbf{L}} = \hat{\mathbf{D}}^{-\frac{1}{2}} \hat{\mathbf{A}} \hat{\mathbf{D}}^{-\frac{1}{2}},
\end{equation}
where $\hat{\mathbf{A}} = \mathbf{A} + \mathbf{I}_d$ is the adjacency matrix with self-loops, $\hat{\mathbf{D}}$ is the corresponding degree matrix with $\hat{\mathbf{D}}_{ii} = \sum_{j=1}^d \hat{\mathbf{A}}_{ij}$, and $\mathbf{I}_d$ is the identity matrix.

The matrix $\hat{\mathbf{L}}$ is a normalized version of $\mathbf{L}$, which captures the graph's structure and enables the smoothing of node features. By incorporating self-loops into the adjacency matrix, we ensure that each node's feature is considered during the smoothing process, enhancing the stability and expressiveness of the learned representations.
Afterwards, the propagation rule in Equation [\ref{formula:GCNpropagation}] of GCN can be modified as follows:

\begin{equation}
\mathbf{Z}^{(l+1)} = \sigma(\hat{\mathbf{L}} \mathbf{Z}^{(l)} \mathbf{W}^{(l)}),
\label{eq:propagationgcn}
\end{equation}

The term $\hat{\mathbf{L}} \mathbf{Z}^{(l)}$ effectively applies the GLSPR to the node features. By multiplying the node features with the normalized GLm $\hat{\mathbf{L}}$, we achieve a smoothing effect that takes into account the graph's structure. This operation allows the model to leverage the connectivity information encoded in the graph to refine the node features and suppress high-frequency noise.

It is important to note that although we do not explicitly compute the eigendecomposition of $\hat{\mathbf{L}}$ in our implementation, the matrix $\hat{\mathbf{L}}$ inherently possesses the spectral properties of the Graph Laplacian.

To examine the effectiveness of our proposed GLSPR, we provide a theoretical convergence analysis under simplified conditions in the Appendix and conduct the ablation study and empirical analyses (see Sec. 4.3). We can also examine the convergence and generalization properties of GLSPR. Let $\mathbf{Z}^*$ denote the optimal node features that minimize the loss function $\mathcal{L}(\mathbf{Z})$. We can show that by incorporating the GLSPR term $\hat{\mathbf{L}} \mathbf{Z}^{(l)}$ into the GCN propagation rule, the learned node features $\mathbf{Z}^{(l)}$ converge to $\mathbf{Z}^*$ under mild assumptions on the graph structure and the loss function. Specifically, if the graph is connected and the loss function is convex and smooth, the iterative updates of the node features using Equation (\ref{eq:propagationgcn}) will converge to the optimal solution $\mathbf{Z}^*$ (see the Appendix).

Finally, the output features can be obtained by passing the final layer's features $\mathbf{Z}^{(L)}$ through a fully connected layer (FC):
\begin{equation}
\mathbf{Z} = \sigma(\bW_{\text{out}}\mathbf{Z}^{(L)} \mathbf{W}^{(L)})
\end{equation}
where $\bW_\text{out} \in \mathbb{R}^{g_\text{dim}\times n_\text{out}}$ denotes the weight of the FC, $n_\text{out}$ and $g_\text{dim}$ stand for the number of neurons of the FC and the embedding dimension of GCN. Finally, the predicted result could be done via

\begin{equation}
\mathbf{\hat{Y}} = \text{Softmax}(\bW_{\text{cls}}\bZ)
\end{equation}
where $\bW_\text{cls} \in \mathbb{R}^{n_\text{out}\times n_\text{cls}}$ denotes the weight of the FC, $n_\text{cls}$ stands for the number of classes, $\hat{\mathbf{Y}}$ is the predicted result of the proposed GRACE for robust DeepFake video detection.

%

\subsection{Sparsity Constraint on Network Optimization}

While the affinity matrix $\mathbf{X}_{\text{FE}}$ effectively captures spatiotemporal dependencies within the feature representation, learning from noisy face sequences remains challenging. To enhance the network's robustness, we introduce the SC as a regularization term in the learning objective, formulated as the $\ell_1$ norm of $\mathbf{X}$, denoted by $|\mathbf{X}|_1$. This constraint encourages the backbone to extract the relatively sparse feature representations from each input frame, thereby reducing the impact of noisy nodes, as illustrated in Figure \ref{fig:features}. Consequently, the model focuses on the most informative features, improving its robustness to noisy face sequences and enhancing its efficacy in DeepFake video detection.

During the training phase, we optimize the model parameters using cross-entropy loss augmented with the SC:

\begin{equation}
\mathcal{L} = -\sum_{c=1}^{n_\text{cls}} \mathbf{\hat{Y}}_{c} \log(\mathbf{Y}_{c}) + \alpha|\mathbf{X}|_1
\end{equation}
where $\mathbf{Y}$ is the ground truth label, and $\alpha$ denotes the weight of the SC, set to $1e^{-5}$ across all experiments in this study.

The synergistic application of GLSPR and SC in our GRACE method yields robust performance in noisy sequences. By incorporating GLSPR into the GCN propagation rule, we enhance the robustness and generalization capability of the learned representations, making our approach particularly suitable for DeepFake video detection in challenging real-world scenarios with noisy and unreliable face sequences. Concurrently, SC guides the network to focus on essential information within $\mathbf{X}_{\text{FE}}$ and suppress irrelevant nodes, while enforcing the learning of sparse features for valid faces.

\section{Experimental Results}\label{sec:experiments}
\subsection{Experimental Configuration}
The robustness validation of the proposed method is the core of our investigation, particularly when applied to noisy face sequences containing many invalid faces. To achieve this, the use of representative benchmark datasets is essential. Therefore, we selected three well-established benchmark datasets for performance evaluation: FF++ \cite{ffplus}, Celeb-DFv2 \cite{celeb}, and the large-scale DFDC dataset \cite{dfdc}. The FF++ dataset \cite{ffplus} comprises four distinct classes of manipulation methods: 1) DeepFakes (DF), 2) Face2Face (F2F), 3) FaceSwap (FS), and 4) NeuralTextures (NT). For each class, a set of 1,000 original videos was used to generate 1,000 manipulated versions, resulting in a total of 1,000 authentic and 4,000 doctored videos. The Celeb-DF dataset \cite{celeb} contains 590 original videos and 5,639 manipulated counterparts, generated using improved generative adversarial networks at a resolution of $256\times 256$. To enhance the quality of the manipulated videos, a Kalman filter is employed in Celeb-DF \cite{celeb} to mitigate temporal inconsistencies between successive frames. The DFDC dataset \cite{dfdc}, created by Facebook in collaboration with other organizations, is a large-scale dataset designed to facilitate the development of DeepFake detection algorithms. It consists of over 100,000 videos, containing a mix of authentic and manipulated content generated using various state-of-the-art face swapping and facial reenactment techniques, ensuring a diverse and challenging set of DeepFakes for evaluation.

\noindent\textbf{Training Hyperparameters of Our GRACE.}
To ensure a balanced performance appraisal, the Celeb-DF \cite{celeb}, FF++ \cite{ffplus}, and DFDC \cite{dfdc} datasets were divided into training, validation, and testing sets following an $8:1:1$ ratio. In line with our objective to ascertain the efficacy of GRACE in the presence of unstable face detectors, we trained separate GRACE models independently on each dataset. During the training phase, the Adam optimizer \cite{kingma2017adam} was utilized with an initial learning rate of $1e^{-4}$ and a step-learning-decay schedule. We employed the 53-layer Cross Stage Partial Network (CSPNet) \cite{bochkovskiy2020yolov4} as the backbone network. Note that any CNNs could be used in our GRACE as backbone network. The standard GCN with our Graph Laplacian was implemented for stacking $g_n$ layers, with $g_n=8$ and embedding size $g_\text{dim}=400$ being the default setting in this study. The number of neurons of the last fully connected layer $n_\text{out}$ is $2048$ for our experiments. All facial images were resized to $144 \times 144$ during both training and inference stages. Standard data augmentation techniques, such as random noise, cropping, and flipping, were adopted during the training phase. The training phase consisted of 200 epochs, with a learning rate decay of 0.1 every 100 epochs. We randomly sampled $N=16$ successive facial images to form the input tensor for our experiments. All comparison methods, including the proposed GRACE, were trained on the training set and evaluated on the testing set.

\noindent\textbf{Training Hyperparameters of Peer Methods.}
For performance evaluation, we compared our proposed method with several state-of-the-art DeepFake detection techniques, including Xception \cite{chollet2017xception}, $F^3$-net \cite{qian2020thinking}, RECCE \cite{RECCE}, DFIL \cite{pan2023dfil}, UCF \cite{yan2023ucf}, CORE \cite{ni2022core}, and TALL-Swin \cite{xu2023tall}. The image-based approaches, namely Xception, $F^3$-net, RECCE, DFIL, UCF, and CORE, were trained using the same strategy as described previously, with their default settings. However, the learning rates of Xception \cite{chollet2017xception} and UCF \cite{yan2023ucf} were adjusted to $2e^{-4}$ for better performance. The video-based approach, TALL-Swin \cite{xu2023tall}, was trained using their default settings. During the training phase, we randomly selected $N$ facial images from the training set. The final authenticity verdict for the input video was determined by averaging the $N$ prediction outcomes corresponding to the $N$ facial images extracted from the input video, using a temporally centered cropping strategy. For all other methods, the number of frames $N$ used was set to $16$. The image size for Xception \cite{chollet2017xception}, $F^3$-net \cite{qian2020thinking}, RECCE \cite{RECCE}, UCF \cite{yan2023ucf}, and CORE \cite{ni2022core} is $256\times 256$, suggested by their default settings, while that for DFIL \cite{pan2023dfil} and TALL-Swin \cite{xu2023tall} are $299\times 299$ and $224\times 224$, respectively. 

\noindent\textbf{Settings in Inference Phase.}
To evaluate the model's performance under the influence of an unstable face detector, we randomly replaced certain facial images with background segments, as determined by the masking ratio $m_r$. We experimented with masking ratios ranging from 0.1 to 0.8 to assess the effectiveness of GRACE under varying levels of noise in the face sequences. For instance, with $N=16$ and $m_r=0.5$, up to eight facial images could be replaced with background images in the corresponding frames, simulating real-world scenarios where face detection may be challenging or unreliable.
In our experimental setup, we sampled $N=16$ frames from the middle portion of each video, following the same approach used during the training process. When $m_r=0.5$, half of the 16 frames (i.e., 8) were randomly replaced with either background or completely black images. By varying the masking ratio, we evaluated the robustness and stability of each method under different levels of noise in the face sequences.

Furthermore, we assumed that each frame should contain at least one face to simulate adversarial attacks on face detectors in real-world scenarios. In cases where no face was detected in a frame, we replaced that frame with a black image, generating a noisy face sequence that allowed us to assess the robustness of GRACE under challenging conditions.
Our experimental analysis employed three performance metrics: accuracy, macro F1-Score, and Area Under the Receiver Operating Characteristic Curve (AUC). 
For simplicity, these metrics are referred to as Accuracy (Acc.), F1-Score, and AUC throughout the experimental sections.

\subsection{Quantitative Results}
\begin{table*}[]
    \centering\small
    \caption{Quantitative comparison of the noisy face sequences under different masking rations $m_r$ between the proposed GRACE and other state-of-the-art methods. We highlight the best performance in red and the second-best performance in blue, considering the several benchmark dataset, such as FF++ \cite{ffplus}, DFDC \cite{dfdc}, and Celeb-DF \cite{celeb} with different $m_r$. 
}
    \scalebox{0.95}{
    \begin{tabular}{c|c|c|c|ccc|ccc|ccc}
    \hline
        \hline
        \multirow{2}{*}{Method}         & \multirow{2}{*}{Venue} &  \multirow{2}{*}{Type} &  \multirow{2}{*}{$m_r$} & \multicolumn{3}{c}{FF++ \cite{ffplus}} & \multicolumn{3}{c}{Celeb-DF \cite{celeb}} & \multicolumn{3}{c}{DFDC \cite{dfdc}} \\ \cline{5-13} 
                                  &                                   &     &       & Acc. & F1 & AUC     & Acc.   & F1 & AUC        & Acc.  & F1 & AUC       \\ \hline
        \multirow{3}{*}{Xception \cite{chollet2017xception}} & \multirow{3}{*}{CVPR2017} & \multirow{3}{*}{Image-based} & 0.0                & 0.925                   & 0.894                 &  0.972                   &  0.861                   & 0.806                   &  0.910                   &  0.953                         &  0.910                  &  0.981                   \\&&
                                                            & 0.4                 & 0.869                   & 0.780                 &  0.871                   &  0.631                   & 0.614                   &  0.782                   & {\blue 0.908}                  & {\blue 0.788}           &  0.866                   \\&&
                                                            & 0.8                 & 0.814                   & 0.594                 &  0.654                   &  0.398                   & 0.389                   &  0.604                   & {\blue 0.864}                  &  0.598                  &  0.647                   \\ \hline
                                                                                  
        \multirow{3}{*}{\text{$F^3$}-net \cite{qian2020thinking}}   & \multirow{3}{*}{ECCV2020} & \multirow{3}{*}{Image-based} & 0.0         & 0.950                   & 0.928                 &  0.986                   & {\blue 0.965}            & {\blue 0.957}           & {\blue 0.993}            &  {\blue 0.957}               &  {\blue0.921}                  & {\blue0.986}                   \\&&
                                                            & 0.4                 & {\blue 0.883}           &  0.798                &  0.888                   &  0.691                   &  0.684                  & {\blue 0.895}            &  0.864                       & 0.655                   &  0.755                   \\&&
                                                            & 0.8                 & 0.818                   &  0.599                &  0.662                   &  0.418                   &  0.407                  & {\blue 0.664}            &  0.850                       &  0.539                  &  0.595                   \\ \hline
                                                                                  
        \multirow{3}{*}{RECCE  \cite{RECCE}}                &  \multirow{3}{*}{CVPR2022} & \multirow{3}{*}{Image-based} &0.0                 & 0.938                   &  0.911                &  0.979                   &  0.941                   &  0.925                  &  0.985                   &  0.940                         &  0.872                  &  0.973                   \\&&
                                                            & 0.4                 & 0.878                   &  0.790                &  0.874                   &  0.678                   &  0.669                  &  0.869                   &  0.900                         &  0.752                  &  0.863                   \\&&
                                                            & 0.8                 & 0.817                   &  0.599                &  0.655                   &  0.414                   &  0.404                  &  0.648                   &  0.861                         &  0.579                  &  0.648                   \\ \hline
        \multirow{3}{*}{CORE \cite{ni2022core}}             & \multirow{3}{*}{CVPRW2022} & \multirow{3}{*}{Image-based} & 0.0                 & 0.948                   &  0.925                &  0.984                   &  0.953                   &  0.940                  &  0.989                   &  0.950                         &  0.903                  &  0.977                   \\&&
                                                            & 0.4                 & 0.883                   &  0.799                &  0.888                   & {\blue 0.858}            & {\blue 0.790}           & { 0.890}                 & { 0.907}                       &  0.781                  & {\blue 0.870}            \\&&
                                                            & 0.8                 & 0.818                   &  0.601                &  0.663                   & {\blue 0.764}            & { 0.572}                & { 0.661}                 & { 0.863}                       & { 0.595}                & {\blue 0.651}            \\ \hline                                                                                  
    \multirow{3}{*}{UCF \cite{yan2023ucf}}                  &  \multirow{3}{*}{CVPR2023} & \multirow{3}{*}{Image-based} &0.0                 & 0.937                   &  0.911                &  0.982                   &  0.856                   &  0.792                  &  0.891                   &  0.890                         &  0.815                  &  0.939                   \\&&
                                                            & 0.4                 & 0.875                   &  0.790                &  0.882                   &  0.626                   &  0.607                  &  0.642                   &  0.871                         &  0.733                  &  0.812                   \\&&
                                                            & 0.8                 & 0.815                   &  0.598                &  0.660                   &  0.397                   &  0.389                  &  0.516                   &  0.851                         &  0.586                  &  0.620                   \\ \hline

        \multirow{3}{*}{DFIL \cite{pan2023dfil}}            & \multirow{3}{*}{ACMMM2023} & \multirow{3}{*}{Image-based} & 0.0                 & {\blue 0.954}           & {\blue 0.939}         & {\blue 0.987}            &  0.957                   &  0.954                  &  0.964                   &  0.940                         &  0.881                  &  0.955                   \\&&
                                                            & 0.4                 &  0.876                  & {\blue 0.808}         & {\blue 0.893}            &  0.695                   & 0.684                   &  0.825                   &  0.886                         &  0.720                  &  0.813                   \\&&
                                                            & 0.8                 &  0.759                  &  0.603                & {\blue 0.665}            &  0.518                   &  0.350                  &  0.644                   &  0.855                         &  0.565                  &  0.621                   \\ \hline
                                                                                  
        \multirow{3}{*}{TALL-Swin \cite{xu2023tall}}        & \multirow{3}{*}{ICCV2023} & \multirow{3}{*}{Video-based} & 0.0                 & 0.913                   &  0.868                &  0.881                   & 0.913                    &  0.933                  &  0.924                   & 0.911                          &  0.812                  & { 0.984}                 \\&&
                                                            & 0.4                 & 0.867                   &  0.767                &  0.740                   & 0.847                    &  0.789                  &  0.825                   &  0.872                         &  0.758                  & { 0.786}                 \\&&
                                                            & 0.8                 & {\blue 0.827}           & {\blue 0.605}         &  0.589                   &  0.745                   & {\blue 0.680}           &  0.645                   &  0.845                         & {\blue 0.688}           &  0.650                   \\ \hline
                                                                                  
        \multirow{3}{*}{DFGaze \cite{DFGAZE}}            & \multirow{3}{*}{TIFS2024} & \multirow{3}{*}{Video-based} & 0.0                 & 0.946           &  0.926   &      0.986            &  0.956                   &  0.954                  &  0.972                   &  0.915                         &  0.881                  &  0.968                   \\&&
                                                            & 0.4                 &  0.854                  & 0.724         & 0.795            &  0.756                  & 0.726                   &  0.824                   &  0.818                         &  0.709                  &  0.743                  \\&&
                                                            & 0.8                 &  0.785                  &  0.652                &  0.656           &  0.612                   &  0.669                  &  0.659                   &  0.798                         &  0.648                  &  0.596                   \\ \hline\hline
                                                                                  
        \multirow{3}{*}{GRACE [Ours]}                        &  \multirow{3}{*}{-} & \multirow{3}{*}{Video-based} &0.0                 & {\red 0.962}            & {\red 0.942}          & {\red 0.989}             & {\red 0.989}             & {\red 0.968}            & {\red 0.998}             & {\red 0.969}                   & {\red 0.942}            & {\red 0.988}             \\&&
                                                            & 0.4                 & {\red 0.958}            & {\red 0.936}          & {\red 0.987}             & {\red 0.970}             & {\red 0.920}            & {\red 0.998}             & {\red 0.969}                   & {\red 0.940}            & {\red 0.988}             \\&&
                                                            & 0.8                 & {\red 0.944}            & {\red 0.916}          & {\red 0.983}             & {\red 0.857}             & {\red 0.738}            & {\red 0.980}             & {\red 0.962}                   & {\red 0.925}            & {\red 0.979}             \\  \hline
        \hline
    \end{tabular}\label{tab:main}}
\end{table*}

\begin{table}[ht]
\centering
\caption{Complexity comparison of different methods in terms of FLOPs , MACs, and \#Params.}
\label{tab:method_comparison}
\scalebox{1.12}{
\begin{tabular}{c|c|c|c}
\hline\hline
Method & FLOPs (T) & MACs (T) & \#Params (M) \\ \hline
Xception \cite{chollet2017xception} & 60.796 & 30.356 & 21.861 \\
$F^3$-net \cite{fakefreq} & 192.604 & 95.880 & 22.125 \\
RECCE \cite{RECCE} & 81.655 & 40.667 & 47.693 \\
CORE \cite{ni2022core} & 60.978 & 30.356 & 21.861 \\
UCF \cite{yan2023ucf} & 180.738 & 90.087 & 46.838 \\
DFIL \cite{pan2023dfil} & 60.976 & 30.356 & 20.811 \\
TALL-Swin \cite{xu2023tall} & 30.318 & 15.125 & 86.920 \\
DFGaze \cite{DFGAZE} & 22.647 & 11.241 & 123.185 \\\hline
GRACE [Ours] & 70.751 & 35.246 & 29.661 \\ \hline\hline
\end{tabular}}\label{tab:complexity_analysis}
\end{table}

The primary performance assessment comparing the handling of invalid facial images between our proposed model, GRACE, and various state-of-the-art schemes is provided in Table \ref{tab:main}. Under optimal conditions, where most facial images are valid, GRACE exhibits competitive results, holding its own against other cutting-edge DeepFake video detection methods such as Xception \cite{chollet2017xception}, MesoNet \cite{afchar2018mesonet}, $F^3$-Net \cite{qian2020thinking}, RECCE \cite{RECCE}, CORE \cite{ni2022core}, DFIL \cite{pan2023dfil}, TALL-Swin \cite{xu2023tall}, and DFGaze \cite{DFGAZE}. It is worth noting that TALL-Swin and DFGaze are a video-based approaches.

Specifically, the F1-Score of GRACE 
slightly surpasses those of its contemporaries under clean cases (i.e., $m_r=0$). This outcome implies that the proposed GRACE with Graph Laplacian Smoothing Prior Regularization is effective and reliable for DeepFake video detection. However, in scenarios where partial face images are invalid due to purposeful attacks on face detectors, the performance of traditional image-based methods, including Xception \cite{chollet2017xception}, MesoNet \cite{afchar2018mesonet}, $F^3$-Net \cite{qian2020thinking}, RECCE \cite{RECCE}, CORE \cite{ni2022core}, UCF \cite{yan2023ucf}, and DFIL \cite{pan2023dfil}, may substantially deteriorate since they fail to consider noisy face sequences in real-world scenarios.

Similarly, the video-level DeepFake detection methods, TALL-Swin \cite{xu2023tall} and DFGaze \cite{DFGAZE}, which heavily rely on temporal cues, may suffer further performance degradation when the masking ratio increases. Invalid faces can cause landmark detection failures and incorrect temporal trajectories. Consequently, the F1-Score of TALL-Swin and DFGaze under a masking ratio of 0.8 in the testing phase are lower than 0.7, implying that all predictions would be categorized as either entirely fake or real. Likewise, the performance of another state-of-the-art video-based DeepFake detection method, TALL-Swin \cite{xu2023tall} and DFGaze \cite{DFGAZE}, are poor when $m_r$ is increased. In stark contrast, all quality indices of our proposed GRACE, evaluated on different datasets, display promising results, suggesting that GRACE is robust and reliable even under highly noisy face sequences (e.g., when $m_r=0.8$). Remarkably, since most DeepFake detection methods fail to discuss the impact of unreliable face sequences, the degraded performance is most likely predictable.

To further demonstrate the efficiency and practicality of the proposed method, we conduct a comprehensive complexity analysis and compare it with other state-of-the-art DeepFake detection methods. Table \ref{tab:complexity_analysis} presents the comparison results in terms of floating-point operations (FLOPs), multiply-accumulate operations (MACs), and the number of model parameters for each method with $16\times 3\times 144\times 144$ tensor for the fair comparison. It is evident that GRACE achieves a remarkable balance between computational complexity and performance. With 70.751 trillion FLOPs, 35.246 trillion MACs, and 29.661 million parameters, GRACE exhibits a moderate computational overhead compared to other methods, such as TALL-Swin \cite{attack1}, DFGaze \cite{DFGAZE}, UCF \cite{yan2023ucf}, and RECCE \cite{RECCE}. Notably, GRACE outperforms these methods in terms of FLOPs and MACs while maintaining a comparable number of parameters. Moreover, GRACE demonstrates superior performance in handling noisy face sequences, as shown in the experimental results, despite having a similar complexity to methods like CORE \cite{ni2022core}, Xception \cite{chollet2017xception}, and DFIL \cite{pan2023dfil}. This highlights the effectiveness of the proposed GRACE in learning discriminative and robust representations for DeepFake detection. The complexity analysis further substantiates GRACE as a practical and efficient solution for real-world DeepFake detection challenges, offering a compelling trade-off between computational resources and detection accuracy.

\begin{figure*}
    \centering
    \subfigure[Comparison of AUC for FF++]{
        \label{fig:result_FF++}
        \includegraphics[width=0.32\textwidth]{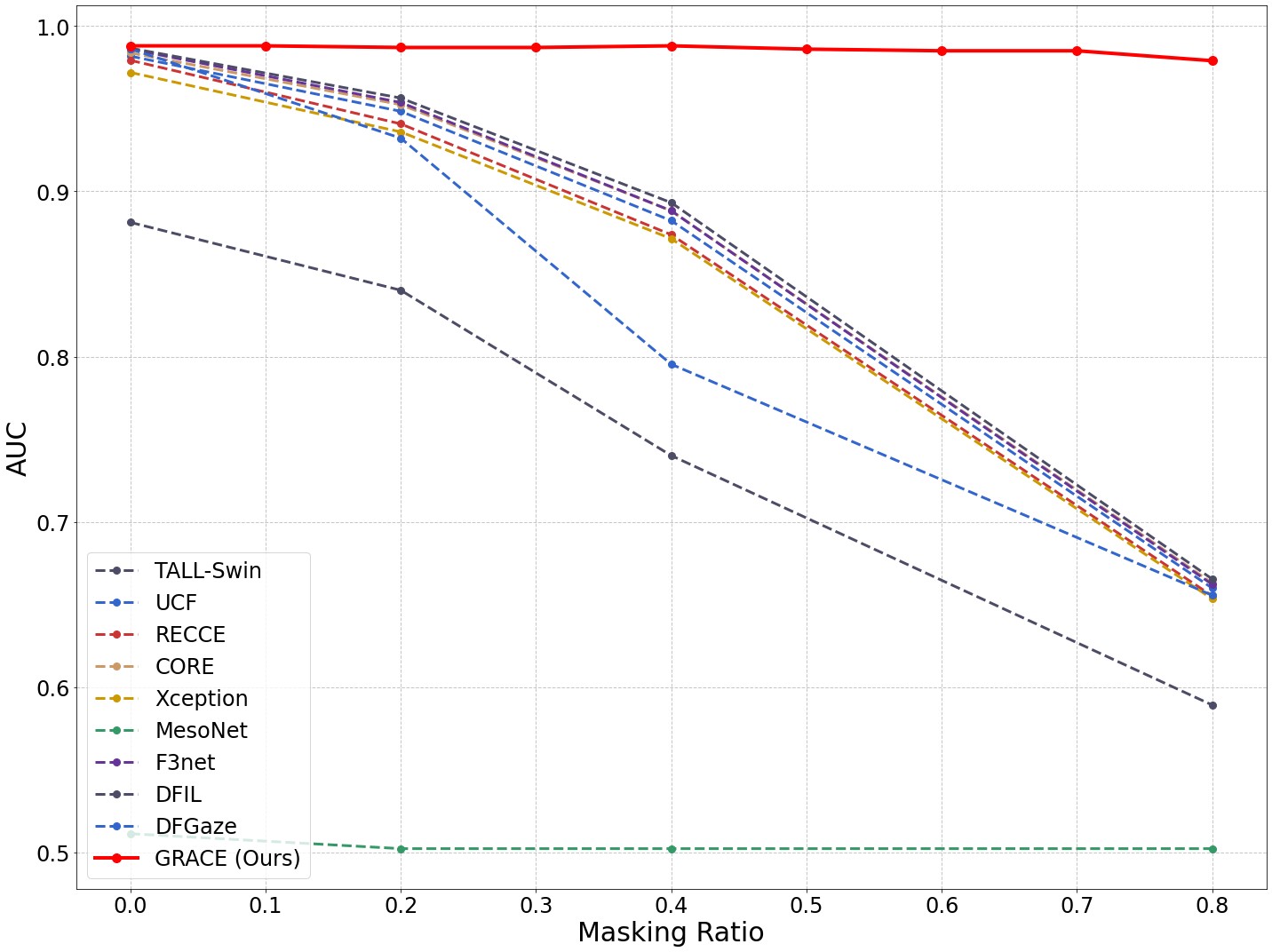}}
    \subfigure[Comparison of AUC for DFDC]{
        \label{fig:result_DFDC}
        \includegraphics[width=0.32\textwidth]{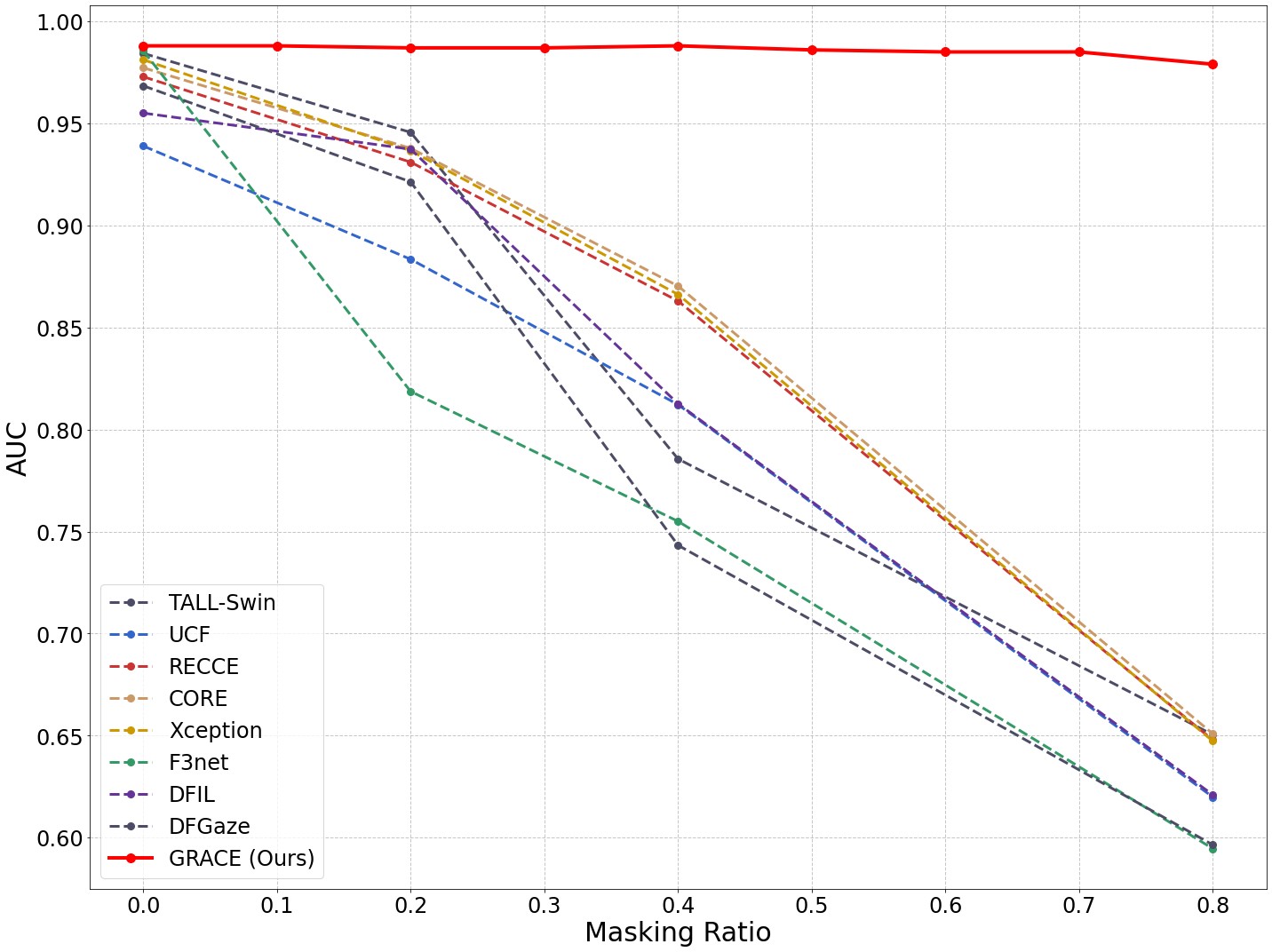}}
    \subfigure[Comparison of AUC for CelebDF]{
        \label{fig:result_celeb}
        \includegraphics[width=0.32\textwidth]{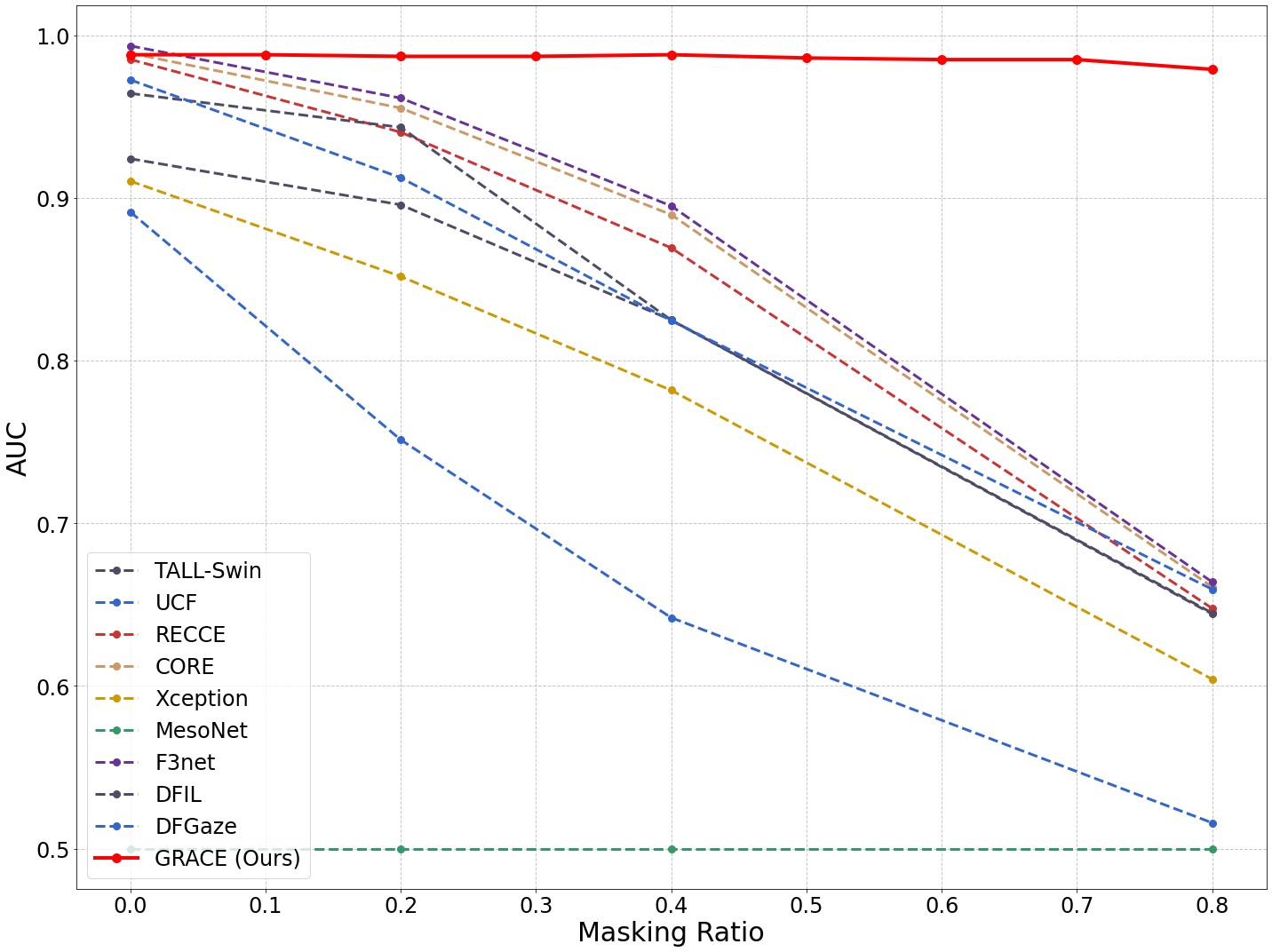}}
    \caption{The performance comparison of the proposed GRACE and other state-of-the-art methods in terms of AUC under different masking ratios $m_r$ for (a) FF++\cite{ffplus}, (b) DFDC \cite{dfdc}, and (c) Celeb-DF \cite{celeb}.}
    \label{Fig.main}
\end{figure*}

The detailed quantitative results, evaluated on benchmark datasets, are illustrated in Figures \ref{fig:result_FF++}, \ref{fig:result_DFDC} and \ref{fig:result_celeb}. In the clean case, i.e., when $m_r=0$, the performance of the proposed method is comparable to other state-of-the-art methods. It is observed that performance degradation becomes increasingly pronounced with a rise in the masking ratio during the testing phase, particularly when the masking ratio ($m_r$) exceeds 0.5. The performance of the previously established TALL-Swin \cite{xu2023tall} and DFGaze \cite{DFGAZE} also decline when the masking ratio surpasses 0.2. A similar trend is discernible in Figure \ref{fig:result_celeb}, which evaluates the DFDC testing set. The performance of contemporary methods diminishes at higher masking ratios, whereas the proposed GRACE method maintains relatively high performance even at a masking ratio of 0.8.
We also draw the AUC comparison between the proposed GRACE and other peer methods in Figure \ref{Fig.main}. We show that the proposed GRACE significantly outperforms other state-of-the-art DeepFake detectors, especially under noisy and unreliable face sequences.

More specifically, most existing DeepFake video/image detection algorithms do not address the impact of noisy face sequences. Although state-of-the-art face detectors perform exceptionally well under pristine conditions, their performance can be severely undermined when subjected to well-engineered post-processing techniques, particularly adversarial perturbations targeting the face detector. Our GRACE method successfully overcomes this shortcoming and introduces a novel and robust DeepFake video detection approach for real-world challenges.

\subsection{Ablation Study}

\begin{figure}
		\centering
		\includegraphics[width=0.5\textwidth]{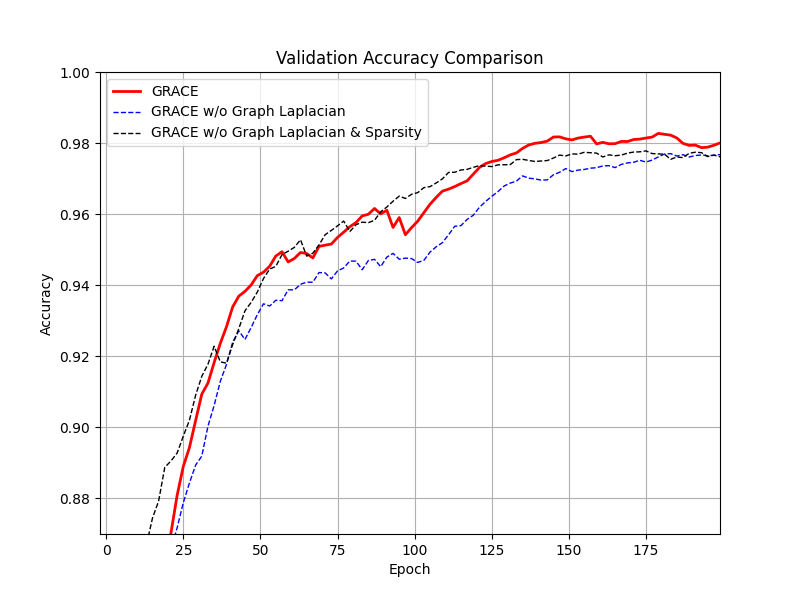}
		\caption{The validation accuracy curve evaluated on FF++ \cite{ffplus} of the proposed GRACE with and without Graph Laplacian Smoothing Prior Regularization and Sparsity Constraint.}
		\label{fig:gcn_curve}	
\end{figure}

Table \ref{tab:module} presents an ablation study for the proposed modules in our GRACE, \textit{i.e.}, GCN, GLSPR, and SC, where the performance is evaluated in noisy face sequences (say, $m_r=0.8$ and $m_r=0.7$). Note that when none of the proposed modules is adopted, we adopt the Transformer \cite{vit} as the classification head with four-head multi-head self-attention (MHSA) with the embedding size of 512 to meet a similar number of parameters with that of our GRACE, which could be treated as a variant of Convolutional Transformer. When we enable the GCN for the proposed feature entanglement and its affinity matrix, the performance of the DeepFake video detection under noisy face sequence, implying that the feature entanglement and its graph representation judiciously embeds the different spatiotemporal features into every node, thereby reducing the impact of invalid faces under noisy face sequences. Furthermore, the proposed GLSPR could improve the robustness since it could filter noisy nodes that might contain many invalid faces without significantly increasing computational complexity. As shown in Figure \ref{fig:gcn_curve}, the convergence of the proposed GRACE with GLSPR remains stable and shows outstanding performance on the FF++ \cite{ffplus} validation set with $m_r=0$, {\red demonstrating that the proposed GRACE still remains training smoothness and increased stably. Moreover, the proposed GRACE shows the better performance on the noisy face sequences, making it more practical.} Finally, the proposed SC significantly benefits the robustness of the DeepFake detection since we could enforce our work learning essential and sparse features for the valid faces. 

\begin{table}
\caption{Ablation study of the proposed GRACE using different classification heads and components. 
}
\centering
\scalebox{0.9}{
\begin{tabular}{c|ccc|ccc|cc}
\hline\hline
$m_r$ & GCN & GLSPR & SC & Acc. & F1 & AUC & \multicolumn{2}{c}{ \#Param / MACs} \\ \hline 
0.8 & & & & 0.844 & 0.655 & 0.705 & \multicolumn{2}{c}{40.79M / 38.82} \\\hline
0.8 & \checkmark & & & 0.926 & 0.873 & 0.931 & \multicolumn{2}{c}{\multirow{4}{*}{29.66M / 35.25}} \\
0.8 & \checkmark & \checkmark & & 0.924 & 0.876 & {\blue 0.977} & & \\
0.8 & \checkmark & & \checkmark & {\red 0.946} & {\blue 0.912} & 0.975 & &\\ 
0.8 & \checkmark & \checkmark & \checkmark & {\blue 0.944} & {\red 0.916} & {\red 0.983}  & &\\ \hline\hline
0.7 & & & & 0.858 & 0.700 & 0.794 & \multicolumn{2}{c}{40.79M / 38.82} \\\hline
0.7 & \checkmark & & & 0.938 & 0.900 & 0.965 & \multicolumn{2}{c}{\multirow{4}{*}{29.66M / 35.25}} \\
0.7 & \checkmark & \checkmark & & 0.950 & 0.920 & {\blue 0.982} & &\\
0.7 & \checkmark & & \checkmark & {\blue 0.952} & {\blue 0.925} & 0.974 & &\\ 
0.7 & \checkmark & \checkmark & \checkmark & {\red 0.960} & {\red 0.938} & {\red 0.985} & &\\ \hline\hline
\end{tabular}\label{tab:module}
}
\end{table}

\begin{figure}
		\centering
		\includegraphics[width=0.5\textwidth]{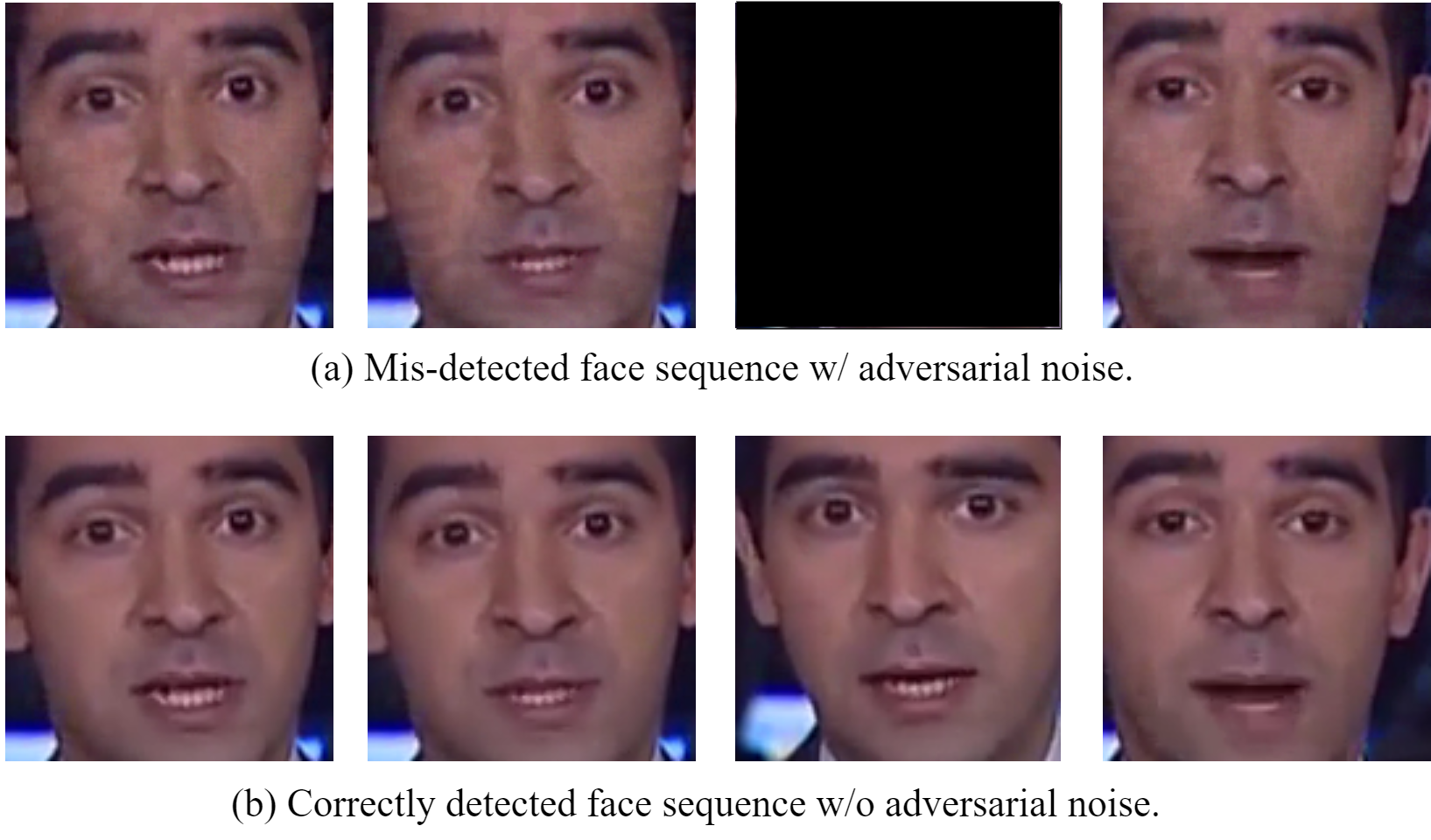}
		\caption{(a) An example of the noisy face sequence caused by PGD-like adversarial attack with $\epsilon=0.04$, $\alpha_\text{atk}=0.01$, and $s=10$, resulting in approximated $m_r=0.2$ condition, where mis-detected faces were replaced with black ones. (b) The corresponding face sequence without adversarial perturbation.}
		\label{fig:adv_atk}	
\end{figure}

\subsection{Adversarial Attack on Face Detector}
As previously discussed, DeepFake videos can be intentionally perturbed to evade detection by face detectors, rendering DeepFake detection ineffective. To emulate this real-world challenge, we employ an open-source adversarial attack on the MTCNN face detector \cite{mtcnn}, leveraging a PGD-like algorithm \cite{carlini2017towards} with a maximum perturbation value of $\epsilon=0.04$, step size $\alpha_\text{adv}=0.01$, and the number of iterations $s=10$ to perturb the test sets of FF++ \cite{ffplus}. The step size $\alpha_\text{adv}$ determines the magnitude of each perturbation step applied to the input image during the iterative adversarial attack process. Assuming that each frame must contain at least one face detectable by MTCNN to assess the rate of missed detections, a black image will replace the frame when no face is detected. The results reveal that the average number of missed detected faces is 3.58 for the FF++ \cite{ffplus} test set, closely mirroring an $m_r=0.2$ scenario.
However, the perturbed face sequences not only make it more difficult to recognize whether the video is fake due to the adversarial noise affecting the image quality, as exemplified in Figure \ref{fig:adv_atk}, but also introduce another challenge: the adversarial attack could cause the face detector to extract non-facial regions (\textit{e.g.}, background). This implies that the actual $m_r$ in this case could be even higher than the estimated $0.2$, as some of the detected faces might not be genuine facial regions. Despite these challenging conditions, our method maintains strong performance compared to other state-of-the-art methods, as illustrated in Table \ref{tab:realworld}. Remarkably, the performance of all peer methods dropped significantly, partly due to the missed detections and partly because the adversarial noise introduces spatial distortions in the facial images. This real-world simulation further substantiates GRACE as a generalized, robust, and effective DeepFake detection model capable of handling noisy face sequences.
\begin{table}
    \caption{The performance comparison of the proposed GRACE and other methods trained on FF++ under simulated real-world scenarios (i.e., adversarial attack on face detector).}
    \centering\small
    \begin{tabular}{c|ccc}
    \hline\hline
    &   \multicolumn{3}{c}{FF++ \cite{ffplus}}  
    \\ \cline{2-4} 
   \multirow{-2}{*}{Method} & Acc. & F1 & AUC 
    \\ \hline
{Xception \cite{chollet2017xception}}           & 0.614 & 0.565 & 0.723  \\
{\text{$F^3$}-net \cite{qian2020thinking}}      & 0.698 & 0.601 & 0.754    \\
{RECCE \cite{RECCE}}                            & 0.760 & 0.633 & 0.824
\\
{CORE \cite{ni2022core}}                        & 0.712 & 0.590 & 0.754  \\ 
{UCF \cite{yan2023ucf}}                         & 0.745 & 0.613 & 0.811  \\ 
{DFIL \cite{pan2023dfil}}                       & 0.755 & 0.691 & 0.805  \\ 
{TALL-Swin \cite{xu2023tall}}                   & 0.796. & 0.715 & 0.848  \\ 
{DFGaze \cite{DFGAZE}}                   & 0.707 & 0.628 & 0.746  \\

    \hline
 GRACE (Ours)                             & {\color{red} 0.910} & {\color{red} 0.883} & {\color{red}  0.937}   \\\hline\hline
    \end{tabular}\label{tab:realworld}
\end{table}

\subsection{Limitations and Discussion}

This study introduces a novel approach, GRACE, to address the challenge of DeepFake video detection in the presence of noisy face sequences. GRACE leverages feature entanglement with sparse constraints and a graph convolutional network with graph Laplacian smoothing prior regularization to effectively exploit the spatial-temporal correlations in face sequences while suppressing the impact of noise and distortions. The experimental results demonstrate the efficacy of GRACE in handling noisy face sequences and achieving state-of-the-art performance on several benchmark datasets.

However, it is essential to acknowledge the limitations of the current study and discuss potential future directions. One limitation is that while our work has shown strong performance on the evaluated datasets, its effectiveness on cross-dataset scenarios, where the training and testing data come from different sources, has not been extensively explored. The robustness of GRACE to domain shifts and variations in noise characteristics across different datasets requires further investigation. Nonetheless, it is worth emphasizing that GRACE represents the first dedicated effort to tackle the problem of noisy face sequences in DeepFake video detection, which has been largely overlooked in previous research. The proposed methodology and insights from this study lay a solid foundation for future work in this important direction.

Another aspect to consider is that GRACE currently does not incorporate masked learning strategies, which have shown promise in handling occlusions and missing data in various computer vision tasks. Integrating masked learning techniques into the GRACE framework could potentially further enhance its robustness to partial occlusions and incomplete face sequences. Moreover, the use of graph convolutional networks in GRACE allows for flexible processing of video frames, as the input frames are not required to be strictly sequential. This property could be leveraged to develop more efficient and adaptive sampling strategies for processing long video sequences.

It is also worth noting that while GRACE has demonstrated significant improvements over existing methods, there is still room for further enhancements. One direction could be to explore more advanced graph neural network architectures, such as graph attention networks or graph transformers, to better capture the complex dependencies and interactions among the spatial-temporal features. Additionally, incorporating prior knowledge or constraints specific to the DeepFake detection domain, such as the consistency of facial landmarks or the coherence of audio-visual signals, could potentially boost the performance and generalizability of the proposed approach.

In conclusion, GRACE represents a significant step forward in addressing the challenge of DeepFake video detection in the presence of noisy face sequences. While acknowledging the limitations and potential areas for improvement, we believe that the proposed methodology opens up new avenues for research in this critical domain. Future work could focus on extending GRACE to handle cross-dataset scenarios, integrating masked learning techniques, exploring more advanced graph neural network architectures, and incorporating domain-specific prior knowledge. As DeepFake techniques continue to evolve and become more sophisticated, developing robust and reliable detection methods that can operate effectively in real-world scenarios with noisy and challenging data remains an ongoing research endeavor of paramount importance.

\section{Conclusions}\label{sec:conclusions}
In this work, we proposed a robust and generalized Graph-Regularized Attentive Convolutional Entanglement approach for DeepFake video detection, specifically designed to address the challenges posed by noisy and unreliable face sequences. The proposed framework leverages spatiotemporal feature entanglement, graph convolutional networks, and graph Laplacian smoothing prior regularization to effectively capture discriminative features while mitigating the impact of invalid facial images.
Extensive experiments 
demonstrate the superior performance and robustness of GRACE compared to state-of-the-art methods, especially under noisy face sequences and adversarial attacks. 
It represents a significant step forward in robust and generalized DeepFake video detection under challenging conditions, contributing to the development of more reliable multimedia forensics techniques in the era of DeepFakes.



	\bibliographystyle{IEEEtran}
	\bibliography{sn-bibliography}	

\vspace{-13mm}
\begin{IEEEbiography}[{\includegraphics[width=1in,height=1.25in,clip]{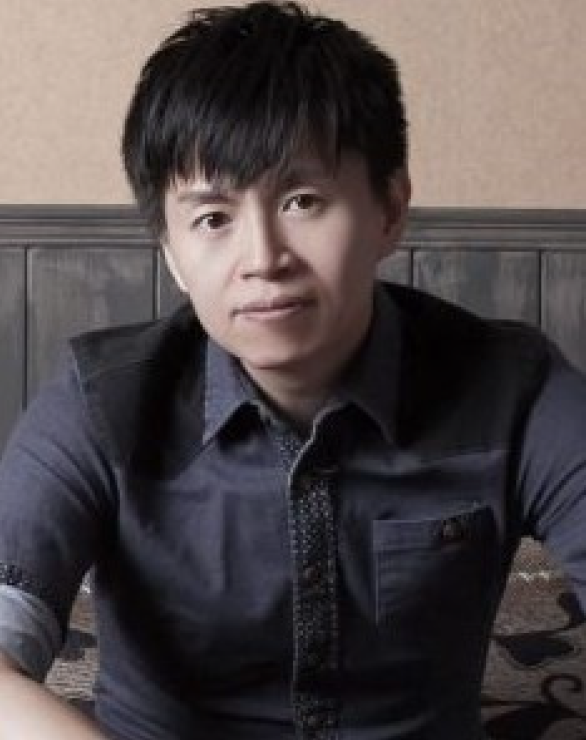}}]%
{Chih-Chung Hsu}
		
(S'11-M'14-SM'20) received his B.S. degree in Information Management from Ling-Tung University of Science and Technology, Taiwan, in 2004, and his M.S. and Ph.D. degrees in Electrical Engineering from National Yunlin University of Science and Technology and National Tsing Hua University (NTHU), Taiwan, in 2007 and 2014, respectively.
From 2014 to 2017, Dr. Hsu was a postdoctoral researcher with the Institute of Communications Engineering at NTHU. He served as an assistant professor with the Department of Management Information Systems at National Pingtung University of Science and Technology from February 2018 to 2021 and has been affiliated with the Institute of Data Science at National Cheng Kung University since 2021. His research interests primarily focus on computer vision and machine/deep learning, with applications in image and video processing. He has published several papers in top-tier journals and conference proceedings, including IEEE TPAMI, IEEE TIP, IEEE TMM, IEEE TGRS, ACM MM, IEEE ICIP, IEEE IGARSS, and IEEE ICASSP.

Dr. Hsu became a Senior Member of the Institute of Electrical and Electronics Engineers (IEEE) in October 2020. He has received multiple awards, including the first-place award of the ACM Multimedia Social Media Prediction Challenge in 2017 and 2019, a top 10\% paper award from the IEEE International Workshop on Multimedia Signal Processing (MMSP) in 2013, and the best student paper award from the IEEE International Conference on Image Processing (ICIP) in 2019. Dr. Hsu leads the Advanced Computer Vision Lab (ACVLab), winning over 20 grand challenges at top-tier conferences over the years, such as the 3rd place award at the Learning to Drive Challenge from the IEEE International Conference on Computer Vision (ICCV), where he was also an invited speaker. He secured the 3rd place award in the Visual Inductive Priors for Data-Efficient Computer Vision Challenge and 1st place in the COV19D challenge at the European Conference on Computer Vision (ECCV) in 2020 and 2022.
In addition to academic research, Dr. Hsu has developed practical research methodologies through his engagement in several industrial-academic projects with organizations like ASE Inc., Leadership Inc., and the Research Center for Advanced Science and Technology. Various domestic media outlets, including PTS, FTV, Taiwan FactCheck Center, United Daily News, and Singapore media have featured his research on DeepFake detection.
\end{IEEEbiography}
\vspace{-14mm}
\begin{IEEEbiography}[{\includegraphics[width=1in,height=1.25in,clip]{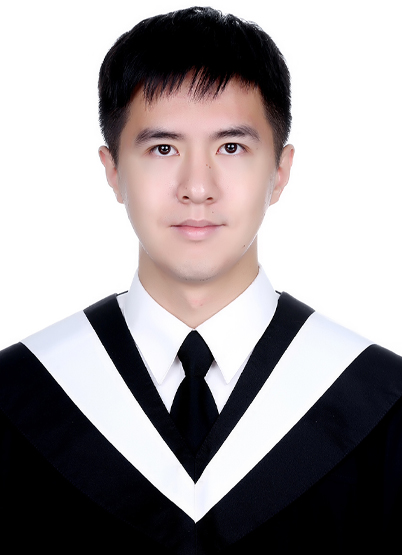}}]%
{Shao-Ning Chen} received his M.S. National Cheng Kung University, Data Science, Taiwan in 2022. His research include image process and computer vision, and focus on   video deepfake detection, especially for unreliable face sequence.
\end{IEEEbiography}
\vspace{-12mm}
\begin{IEEEbiography}[{\includegraphics[width=1in,height=1.25in,clip]{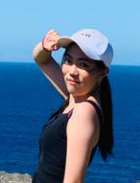}}]%
{Mei-Shuan Wu} received her M.S. National Cheng Kung University, Data Science, Taiwan in 2022. Her research include image process and computer vision, and focus on   video deepfake detection, especially for unreliable face sequence.
\end{IEEEbiography}
\vspace{-12mm}
\begin{IEEEbiography}[{\includegraphics[width=1in,height=1.25in,clip]{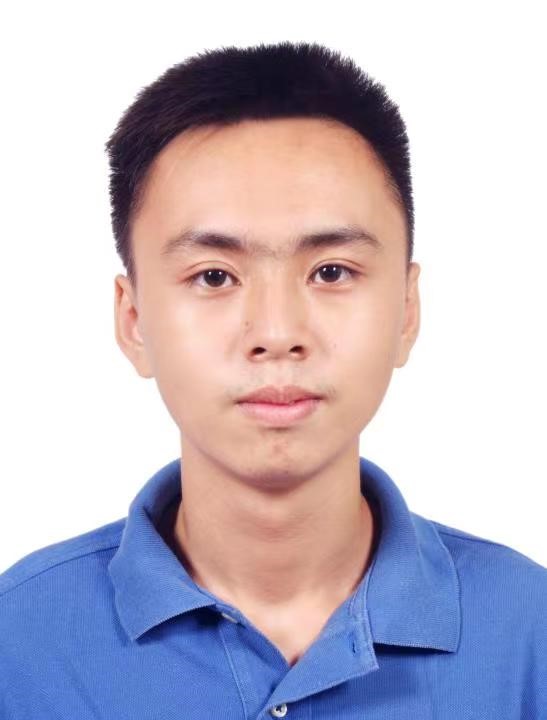}}]%
{Yi-Fan Wang} received his B.S. degree in Data Science and Big Data Analytics from Providence University, Taichung, Taiwan, in 2022. He is currently pursuing an M.S. degree in the institute of Data Science at National Cheng Kung University, Tainan, Taiwan. His research interests include computer vision and deepfake detection.
\end{IEEEbiography}
\vspace{-15mm}
\begin{IEEEbiography}[{\includegraphics[width=1in,height=1.25in,clip]{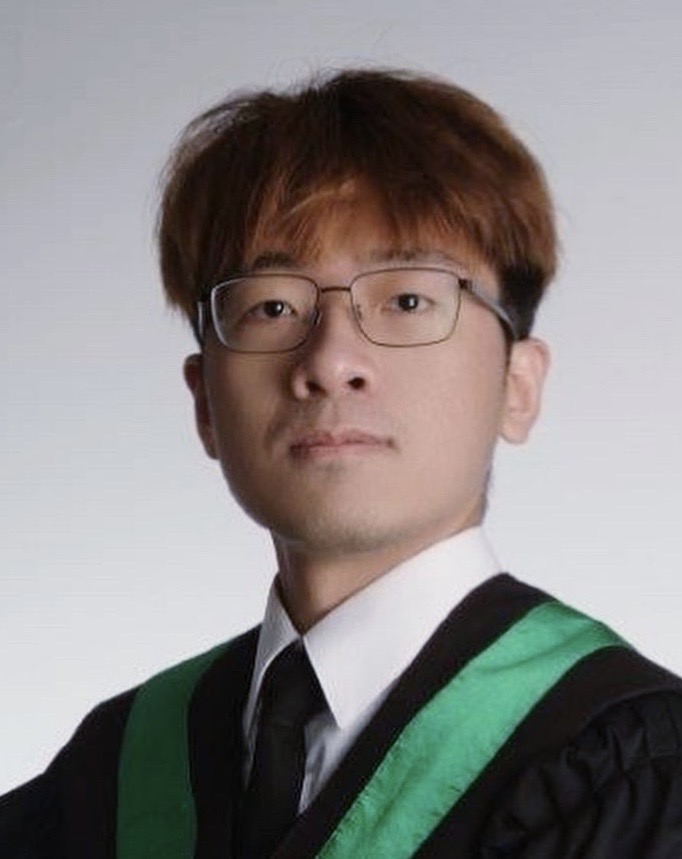}}]%
    {Chia-Ming Lee } received his B.S. degree at the Department of Statistics and Information Science, Fu Jen Catholic University (FJCU), Taiwan, in 2023. He is currently a M.S. student with the Advanced Computer Vision Laboratory (ACVLab) at the Institute of Data Science, National Cheng Kung University (NCKU), Tainan, Taiwan. His research interests mainly focus on image restoration, machine/deep learning with applications in image processing. 
    
    He has received several awards in Top-tier international conference,  including 
    the Jury Prize at Visual Inductive Priors VIP Workshop at ICCV 2023, the 1st place in the AI-MIA Workshop and COVID-19 Diagnosis Competition held at ICASSP 2023, the Top paper award and Top-performance award in the Social Media Prediction SMP Challenge at MM in 2023 and 2024, the 2nd from AutoWCEBleedGen-V2 Challenge at ICIP 2024, the 3rd place in the DEF-AI-MIA Workshop and COVID-19 Detection Challenge associated with CVPR 2024, the Top-3\% performance in the SISR Challenge for New Trends in Image Restoration and Enhancement NTIRE Workshop at CVPR 2024, and the 1st place in Beyond Visible Spectrum Challenge ICPR 2024. 
\end{IEEEbiography}
\vspace{-12.5mm}
\begin{IEEEbiography}[{\includegraphics[width=1in,height=1.25in,clip]{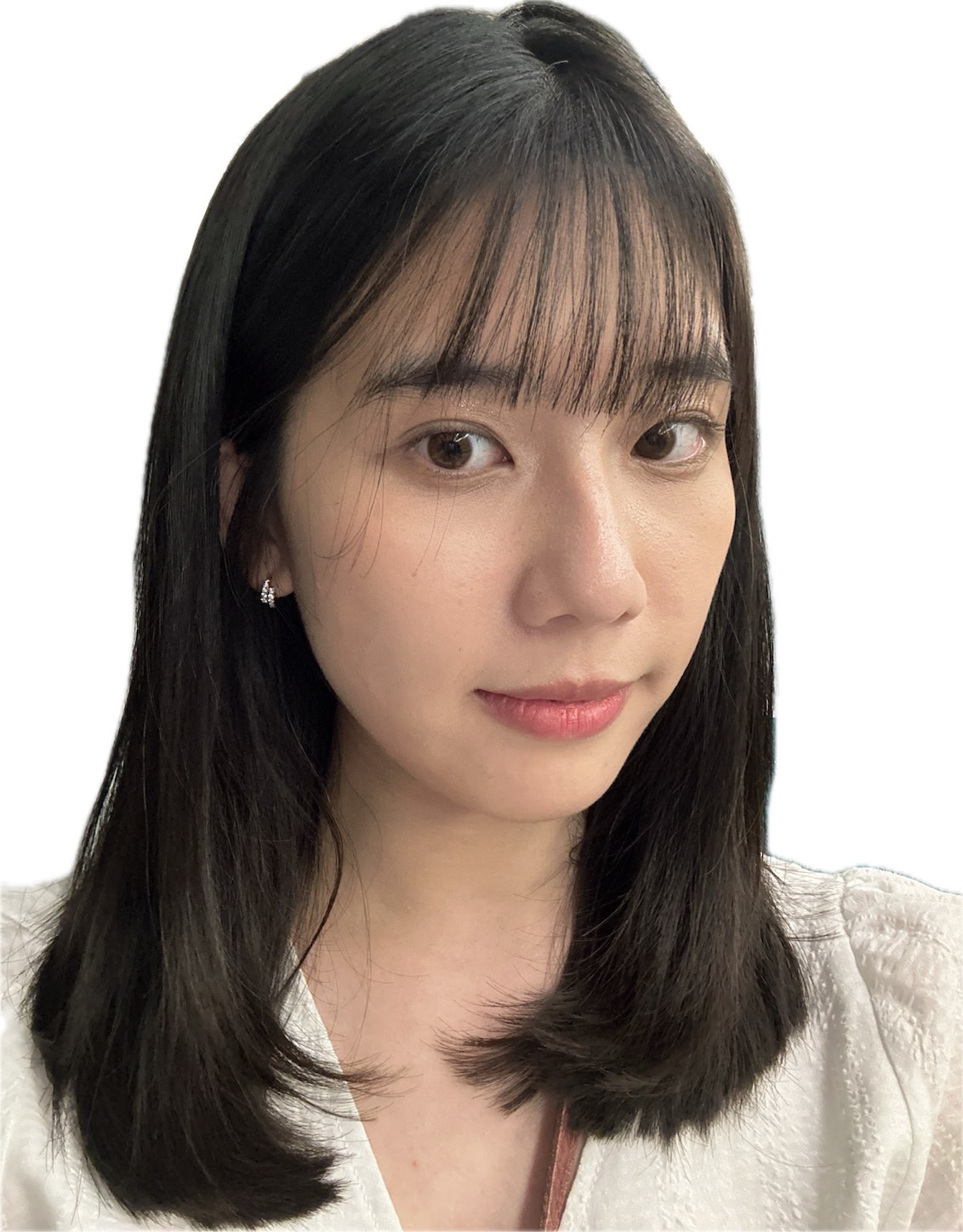}}]%
{Yi-Shiuan Chou} received her B.S. degree at the Department of Statistics, National Cheng Kung University (NCKU), Taiwan, in 2024. She is currently a research assistant in the Advanced Computer Vision Laboratory (ACVLab) at the Institute of Data Science, National Cheng Kung University (NCKU), Tainan, Taiwan. Her research interests mainly focus on computer vision, deep learning and their application.
  
    She secured third place in the COVID-19 Detection Challenge at the DEF-AI-MIA workshop, associated with CVPR 2024. Additionally, she ranked in the top 3\% at the SISR Challenge during the NTIRE workshop at the same conference. Moreover, she earned Top-performance award in the Social Media Prediction (SMP) Challenge at ACM Multimedia (ACMMM) 2024.
\end{IEEEbiography}

\appendix

\title{Appendix of GRACE: Graph-Regularized Attentive Convolutional Entanglement with Laplacian Smoothing for Robust DeepFake Video Detection}


\maketitle
\section*{Theorem.1: Convergence of Graph Laplacian Regularized GCN}
Let $\mathcal{L}(\mathbf{Z})$ be the loss function for a Graph Convolutional Network (GCN) with Graph Laplacian regularization, where $\mathbf{Z} \in \mathbb{R}^{n \times d}$ represents the node features, $n$ is the number of nodes, and $d$ is the feature dimension. Suppose the loss function $\mathcal{L}(\mathbf{Z})$ is convex and smooth in a local neighborhood of the optimal solution $\mathbf{Z}^*$. The GCN propagation rule with Graph Laplacian regularization is given by:

\begin{equation}
\mathbf{Z}^{(l+1)} = \sigma(\hat{\mathbf{L}} \mathbf{Z}^{(l)} \mathbf{W}^{(l)}),
\end{equation}
where $\hat{\mathbf{L}} = \hat{\mathbf{D}}^{-\frac{1}{2}} \hat{\mathbf{A}} \hat{\mathbf{D}}^{-\frac{1}{2}}$ is the normalized Graph Laplacian matrix, $\hat{\mathbf{A}}$ is the adjacency matrix with self-loops, $\hat{\mathbf{D}}$ is the corresponding degree matrix, $\mathbf{W}^{(l)}$ is the weight matrix at layer $l$, and $\sigma$ is a non-expansive activation function satisfying $|\sigma(\mathbf{x}) - \sigma(\mathbf{y})| \leq L_{\sigma}|\mathbf{x} - \mathbf{y}|$ for any $\mathbf{x}, \mathbf{y}$, with Lipschitz constant $L_{\sigma} \leq 1$. If the following conditions hold:
\begin{enumerate}
\item The spectral norm of the normalized Graph Laplacian matrix $|\hat{\mathbf{L}}|$ is bounded by $\lambda_{\max}(\hat{\mathbf{L}}) \leq 2$. The eigenvalues of $\hat{\mathbf{L}}$ lie in the interval $[0, 2]$, with the smallest eigenvalue $\lambda_{\min}(\hat{\mathbf{L}}) = 0$ if and only if the graph is connected.

\item The weight matrices $\mathbf{W}^{(l)}$ have bounded spectral norms, i.e., $|\mathbf{W}^{(l)}| \leq B_W$ for all $l$, and their variations are appropriately constrained \cite{li2018deeper}. This assumption ensures stable learning dynamics and prevents exploding gradients.

\item The composition of functions in each GCN layer, given by $f^{(l)}(\mathbf{Z}) = \sigma(\hat{\mathbf{L}} \mathbf{Z} \mathbf{W}^{(l)})$, is Lipschitz continuous with constant $L_f = L_{\sigma} \cdot \lambda_{\max}(\hat{\mathbf{L}}) \cdot B_W$.
\item The iterates $\mathbf{Z}^{(l)}$ stay within a local neighborhood of the optimal solution $\mathbf{Z}^*$ where the loss function is convex. This assumption is reasonable when the GCN is properly initialized and the learning rate is appropriately chosen.
\end{enumerate}

Then, the iterative updates of the node features using the Graph Laplacian regularized GCN propagation rule will converge to the optimal solution $\mathbf{Z}^*$, \textit{i.e.,}

\begin{equation}
\lim_{l \rightarrow \infty} |\mathbf{Z}^{(l)} - \mathbf{Z}^*| = 0
\end{equation}.

Moreover, the convergence rate is linear, satisfying:
\begin{equation}
|\mathbf{Z}^{(l+1)} - \mathbf{Z}^*| \leq \rho |\mathbf{Z}^{(l)} - \mathbf{Z}^*|,
\end{equation}
where $\rho = L_f < 1$ is the contraction factor.

The proof follows from the convergence analysis of iterative algorithms for convex optimization problems. Consider two consecutive iterations $\mathbf{Z}^{(l)}$ and $\mathbf{Z}^{(l+1)}$:
\begin{align}
|\mathbf{Z}^{(l+1)} - \mathbf{Z}^*| &= |f^{(l)}(\mathbf{Z}^{(l)}) - f^{(l)}(\mathbf{Z}^*)| \\
&\leq L_f |\mathbf{Z}^{(l)} - \mathbf{Z}^*|,
\end{align}
where $L_f = L_{\sigma} \cdot \lambda_{\max}(\hat{\mathbf{L}}) \cdot B_W$ is the Lipschitz constant of the composition of functions in each GCN layer. Under the assumptions that $\mathbf{Z}^{(l)}$ is sufficiently close to $\mathbf{Z}^*$ and the learning dynamics are stable, we have $L_f < 1$, which implies:
\begin{equation}
|\mathbf{Z}^{(l+1)} - \mathbf{Z}^*| \leq \rho |\mathbf{Z}^{(l)} - \mathbf{Z}^*|,
\end{equation}
where $\rho = L_f < 1$ is the contraction factor. This result shows that the iterative sequence ${\mathbf{Z}^{(l)}}$ converges linearly to the optimal solution $\mathbf{Z}^*$ within its local neighborhood \cite{nesterov2018lectures}.

The boundedness of the spectral norm and the eigenvalue properties of the normalized Graph Laplacian matrix $\hat{\mathbf{L}}$ ensure stable propagation of node features and prevent over-smoothing \cite{li2018deeper,oono2019graph}. In the context of DeepFake detection, this implies that the GCN can effectively capture discriminative spatial-temporal patterns while mitigating noise and distortions.
The bounded spectral norms and constrained variations of the weight matrices $\mathbf{W}^{(l)}$ promote stable learning dynamics and prevent exploding gradients \cite{li2018deeper}. This assumption is crucial for training deep GCNs on large-scale DeepFake datasets, where the model needs to learn robust and generalizable representations.
The Lipschitz continuity of the composition of functions in each GCN layer, with a contraction factor $\rho < 1$, guarantees the convergence of the iterative feature updates to the optimal solution. This property ensures that the Graph Laplacian regularized GCN can effectively learn discriminative node embeddings for DeepFake detection, even in the presence of challenging conditions such as partial occlusions, temporal inconsistencies, and diverse manipulation techniques.
In summary, the convergence theorem for Graph Laplacian regularized GCNs, along with its assumptions and practical implications, provides a strong theoretical foundation for the proposed method in the context of DeepFake detection. The theorem justifies the effectiveness of incorporating Graph Laplacian regularization into GCNs for learning robust and discriminative spatial-temporal representations from noisy and challenging DeepFake videos.

\section*{Appendix.1: Hyperparameters Selection}
\begin{table}
    \caption{Performance evaluation of the proposed GRACE with different hyperparameter settings using FF++ \cite{ffplus}. $g_\text{dim}$ and $g_n$ are the embedding dimension and number of layers of GCN, respectively; $N$ is the frames extracted from the video; $n_\text{out}$ is the number of neurons of FC; $\alpha$ stands for weights of sparsity. }
    \centering
\begin{tabular}{c | c  l l |c c c | c  l l |c c c}
\hline \hline 
$m_r$ & \multicolumn{3}{c|}{$N$}         & Acc.      &  F1       & AUC       & \multicolumn{3}{c|}{$g_n$}       & Acc.      & F1         & AUC     \\ \hline \hline
0.8     & \multicolumn{3}{c|}{12}        & 0.922	& 0.875	    & 0.971     & \multicolumn{3}{c|}{12}          & 0.856    & 0.711      & 0.950\\
0.8     & \multicolumn{3}{c|}{20}        & {\red 0.948}	& {\red 0.918}	    & 0.978     & \multicolumn{3}{c|}{4}           & {\red 0.948}    & {\red 0.918}      & 0.982\\
0.8     & \multicolumn{3}{c|}{16}        & 0.944	& 0.916	    & {\red 0.983}     & \multicolumn{3}{c|}{8}           & 0.944	  & 0.916	   & {\red 0.983} \\\hline
0.7     & \multicolumn{3}{c|}{12}        & 0.958	& 0.936	    & 0.983     & \multicolumn{3}{c|}{12}          & 0.896    & 0.816      & 0.964\\
0.7     & \multicolumn{3}{c|}{20}        & 0.954	& 0.930	    & {\red 0.986}     & \multicolumn{3}{c|}{4}           & 0.952    & 0.926      & 0.983\\
0.7     & \multicolumn{3}{c|}{16}        & {\red 0.960}	& {\red 0.938}	    & 0.985     & \multicolumn{3}{c|}{8}           & {\red 0.960}	  & {\red 0.938}	   & {\red 0.985}\\ \hline

-       & \multicolumn{3}{c|}{$\alpha$}  & Acc.      &  F1       & AUC       & \multicolumn{3}{c|}{$g_\text{dim}$}       & Acc.      & F1         & AUC     \\ \hline\hline
0.8     & \multicolumn{3}{c|}{$1e^{-7}$}   & 0.928    & 0.887     & 0.978     & \multicolumn{3}{c|}{600}         & 0.896	  & 0.805	   & 0.966\\
0.8     & \multicolumn{3}{c|}{$1e^{-6}$}   & 0.942    & 0.910     & 0.974     & \multicolumn{3}{c|}{200}         & 0.934	  & 0.886	   & 0.981\\
0.8     & \multicolumn{3}{c|}{$1e^{-5}$}   & {\red 0.944}	& {\red 0.916}	    & {\red 0.983}     & \multicolumn{3}{c|}{400}         & {\red 0.944}	  & {\red 0.916}	   & {\red 0.983}\\\hline
0.7     & \multicolumn{3}{c|}{$1e^{-7}$}   & 0.954    & 0.930     & 0.981     & \multicolumn{3}{c|}{600}         & 0.938	  & 0.896	   & 0.984\\
0.7     & \multicolumn{3}{c|}{$1e^{-6}$}   & 0.944    & 0.914     & 0.979     & \multicolumn{3}{c|}{200}         & 0.942	  & 0.905	   & {\red 0.987}\\
0.7     & \multicolumn{3}{c|}{$1e^{-5}$}   & {\red 0.960}	& {\red 0.938}	    & {\red 0.985}     & \multicolumn{3}{c|}{400}         & {\red 0.960}	  & {\red 0.938}	   & 0.985\\ \hline\hline

-       & \multicolumn{3}{c|}{$n_\text{out}$}  & Acc.      &  F1       & AUC       & \multicolumn{3}{c}{}           & \multicolumn{3}{c}{}    \\ 
0.8     & \multicolumn{3}{c|}{1024}   & 0.926	&0.871	&0.969     & \multicolumn{3}{c}{}           & \multicolumn{3}{c}{}    \\
0.8     & \multicolumn{3}{c|}{3072}   & 0.924	& 0.876	    & 0.939     & \multicolumn{3}{c}{}           & \multicolumn{3}{c}{}    \\
0.8     & \multicolumn{3}{c|}{2048}   & {\red 0.944}	& {\red 0.916}	    & {\red 0.983}     & \multicolumn{3}{c}{}           & \multicolumn{3}{c}{}    \\
0.7     & \multicolumn{3}{c|}{1024}   & 0.944	&0.908	&0.984     & \multicolumn{3}{c}{}           & \multicolumn{3}{c}{}    \\
0.7     & \multicolumn{3}{c|}{3072}   & 0.926	& 0.880	    &  0.968     & \multicolumn{3}{c}{}           & \multicolumn{3}{c}{}    \\
0.7     & \multicolumn{3}{c|}{2048}   & {\red 0.960}	& {\red 0.938}	    & {\red0.985}     & \multicolumn{3}{c}{}           & \multicolumn{3}{c}{}    
\end{tabular}\label{tab:hp_select}
\end{table}

To achieve optimal performance and robustness, we conducted a comprehensive ablation study to investigate the impact of various hyperparameters on the proposed GRACE method. This analysis provides valuable insights into the design choices and trade-offs involved in developing an effective DeepFake video detection system for real-world scenarios with noisy face sequences. Table \ref{tab:hp_select} presents the performance comparison of GRACE under different hyperparameter settings, evaluated on the challenging FF++ dataset 
\cite{ffplus}.

\subsubsection*{1-1: Number of Extracted Frames ($N$)}
The number of frames employed during the training and testing phases is a crucial aspect of GRACE. While using a larger number of frames might intuitively improve performance, it also significantly increases the computational complexity. To strike an optimal balance, we investigated the impact of varying the number of extracted frames. As shown in Table \ref{tab:hp_select}, using $N=8$ frames results in the lowest computational complexity but slightly compromises performance in terms of Accuracy, Macro F1-Score, and AUC. Conversely, increasing the number of frames to $N=20$ achieves state-of-the-art performance for most masking ratios during testing. Considering the trade-off between effectiveness and efficiency, we recommend using $N=16$ frames as the optimal setting for GRACE.

\subsubsection*{1-2: Number of GCN Layers ($g_n$)}
The depth of the Graph Convolutional Network (GCN) plays a vital role in learning robust feature representations. However, stacking too many layers with the Graph Laplacian smooth prior may lead to over-smoothing of nodes and reduce the discriminative power. We explored the impact of varying the number of GCN layers ($g_n$) in GRACE. As presented in Table \ref{tab:hp_select}, setting $g_n=12$ results in suboptimal performance compared to $g_n=8$ and $g_n=4$, likely due to convergence difficulties within the given 200 epochs. While $g_n=4$ achieves outstanding performance overall, it slightly underperforms in highly noisy conditions (i.e., $m_r=0.8$) compared to $g_n=8$. Therefore, we suggest using $g_n=8$ as a balanced choice for stable and robust performance across various noise levels.

\subsubsection*{1-3: Sparsity Penalty Term ($\alpha$)}
The sparsity penalty term $\alpha$ in the proposed loss function controls the balance between the sparsity constraint and the classification objective. A higher value of $\alpha$ encourages GRACE to learn a sparser feature representation, which is particularly beneficial for DeepFake video detection in the presence of invalid facial images. We investigated the impact of $\alpha$ by varying its value from $1e^{-7}$ to $1e^{-5}$. As shown in Table \ref{tab:hp_select}, a higher sparsity penalty enhances the network's ability to learn essential and discriminative features, thereby reducing the influence of invalid faces and improving overall performance. However, setting $\alpha$ higher than $1e^{-5}$ leads to convergence difficulties. Based on our analysis, we recommend using $\alpha=1e^{-5}$ to achieve a balanced trade-off between sparsity and convergence stability.

\subsubsection*{1-4: GCN Embedding Dimension ($g_\text{dim}$)}
The embedding dimension of the GCN ($g_\text{dim}$) determines the richness of the learned feature representations for DeepFake video detection. We investigated the impact of $g_\text{dim}$ by comparing the performance of GRACE with $g_\text{dim} \in {200, 400, 600}$, as shown in Table \ref{tab:hp_select}. Since the dimension of the graph representation $\bA$ is $400\times 400$, intuitively, the best performance is achieved when $g_\text{dim}=400$. Reducing $g_\text{dim}$ below this value limits the expressive power of the GCN, while increasing it beyond introduces redundancy and harms performance. Therefore, we suggest setting $g_\text{dim}=400$ for optimal results.

\subsubsection*{1-5: Number of Fully Connected Layer Neurons ($n_\text{out}$)}
To aggregate the output of the GCN and feed it into the softmax classifier, a simple fully connected (FC) layer is employed, projecting the graph representation to an $n_\text{out}$-dimensional feature vector. We investigated the impact of $n_\text{out}$ by comparing the performance of GRACE with $n_\text{out} \in {1024, 2048, 3072}$, as shown in Table \ref{tab:hp_select}. While $n_\text{out}=2048$ achieves excellent performance under highly noisy face sequences, the performance gap between $n_\text{out}=2048$ and $n_\text{out}=1024$ is insignificant, suggesting that the choice of $n_\text{out}$ is not highly sensitive. Based on our analysis, we recommend setting $n_\text{out}=2048$ for a good balance between performance and computational complexity.

The comprehensive analysis of the hyperparameters presented in this section highlights the robustness and effectiveness of the proposed GRACE method under various hyperparameter settings. By carefully selecting these hyperparameters, GRACE achieves state-of-the-art performance in DeepFake video detection, even in challenging real-world scenarios with noisy face sequences. The insights gained from this analysis provide valuable guidance for practitioners and researchers aiming to develop robust and efficient DeepFake detection systems.

\end{document}